\documentclass[journal]{IEEEtran}
\usepackage{graphics} 
\usepackage{epsfig} 
\usepackage{times} 
\usepackage{amsmath, bm} 
\usepackage{amssymb}  
\usepackage{mathtools}
\usepackage{multirow}
\usepackage{tensor}
\usepackage{graphicx}
\usepackage{arydshln,leftidx,mathtools}
\makeatletter
\let\NAT@parse\undefined
\makeatother
\usepackage{hyperref}
\usepackage{subcaption}
\usepackage{xcolor}      
\newcommand{\overbar}[1]{\mkern 1.5mu\overline{\mkern-1.5mu#1\mkern-1.5mu}\mkern 1.5mu}
\ifCLASSINFOpdf
\else
\fi
\begin{document}
%
\title{\huge Soft Synergies: Model Order Reduction of Hybrid Soft-Rigid Robots via Optimal Strain Parameterization}
%
%
%

\author{Abdulaziz~Y.~Alkayas,~Anup~Teejo~Mathew,~Daniel Feliu-Talegon,~Ping~Deng,\\ ~Thomas~George~Thuruthel~and~Federico~Renda
\thanks{Received 16 October 2024; accepted 3 December 2024. Date of publication 25 December 2024; date of this version 19 February 2025}
\thanks{This work was
supported by the US Office of Naval Research Global under Grant N62909-
21-1-2033 and in part by Khalifa University under Awards No. RIG-2023-048,
RC1-2018-KUCARS.}
\thanks{Abdulaziz Y. Alkayas is with the Department of Mechanical and Nuclear Engineering, Khalifa University, Abu Dhabi, UAE, and also with the Department of Computer Science, University College London, WC1E 6BT London, U.K. (e-mail: 100052628@ku.ac.ae).}
\thanks{Anup Teejo Mathew and Federico Renda arewith the Department of Mechanical and Nuclear Engineering, Khalifa University, Abu Dhabi, UAE, and also with the Khalifa University Center for Autonomous Robotic Systems (KUCARS), Khalifa University, Abu Dhabi, UAE (e-mail: anup.mathew@ku.ac.ae; federico. renda@ku.ac.ae).}%
\thanks{Daniel Feliu-Talegon and Ping Deng are with the Department of Mechanical and Nuclear Engineering, Khalifa University, Abu Dhabi, UAE (e-mail: daniel.talegon@ku.ac.ae; 100064424@ku.ac.ae).}
\thanks{Thomas George Thuruthel is with the Department of Computer Science, University College London, WC1E 6BT London, U.K. (e-mail: t.thuruthel@ucl.ac.uk).} 
\thanks{This article has supplementary downloadable material available at https://doi.org/10.1109/TRO.2024.3522182, provided by the authors.}
\thanks{Digital Object Identifier 10.1109/TRO.2024.3522182}
}

%
%

\markboth{IEEE TRANSACTIONS ON ROBOTCS,~Vol.~41,~2025}%
{Alkayas \MakeLowercase{\textit{et al.}}: Bare Demo of IEEEtran.cls for IEEE Journals}
%



\maketitle

\begin{abstract}
Soft robots offer remarkable adaptability and safety advantages over rigid robots, but modeling their complex, nonlinear dynamics remains challenging. Strain-based models have recently emerged as a promising candidate to describe such systems, however, they tend to be high-dimensional and time-consuming. This paper presents a novel model order reduction approach for soft and hybrid robots by combining strain-based modeling with Proper Orthogonal Decomposition (POD). The method identifies optimal coupled strain basis functions -or mechanical synergies- from simulation data, enabling the description of soft robot configurations with a minimal number of generalized coordinates. The reduced order model (ROM) achieves substantial dimensionality reduction in the configuration space while preserving accuracy. Rigorous testing demonstrates the interpolation and extrapolation capabilities of the ROM for soft manipulators under static and dynamic conditions. The approach is further validated on a snake-like hyper-redundant rigid manipulator and a closed-chain system with soft and rigid components, illustrating its broad applicability. Moreover, the approach is leveraged for shape estimation of a real six-actuator soft manipulator using only two position markers, showcasing its practical utility. Finally, the ROM's dynamic and static behavior is validated experimentally against a parallel hybrid soft-rigid system, highlighting its effectiveness in representing the High-Order Model (HOM) and the real system. This POD-based ROM offers significant computational speed-ups, paving the way for real-time simulation and control of complex soft and hybrid robots.
\end{abstract}

\begin{IEEEkeywords}
Modeling, control, and learning for Soft Robots, Reduced order modeling, Proper orthogonal decomposition, Strain parameterization, Cosserat Rod.
\end{IEEEkeywords}

%
\IEEEpeerreviewmaketitle

\section{Introduction}
%
%
%
%
\begin{figure}[]
    \centering
    \includegraphics[width=\linewidth]{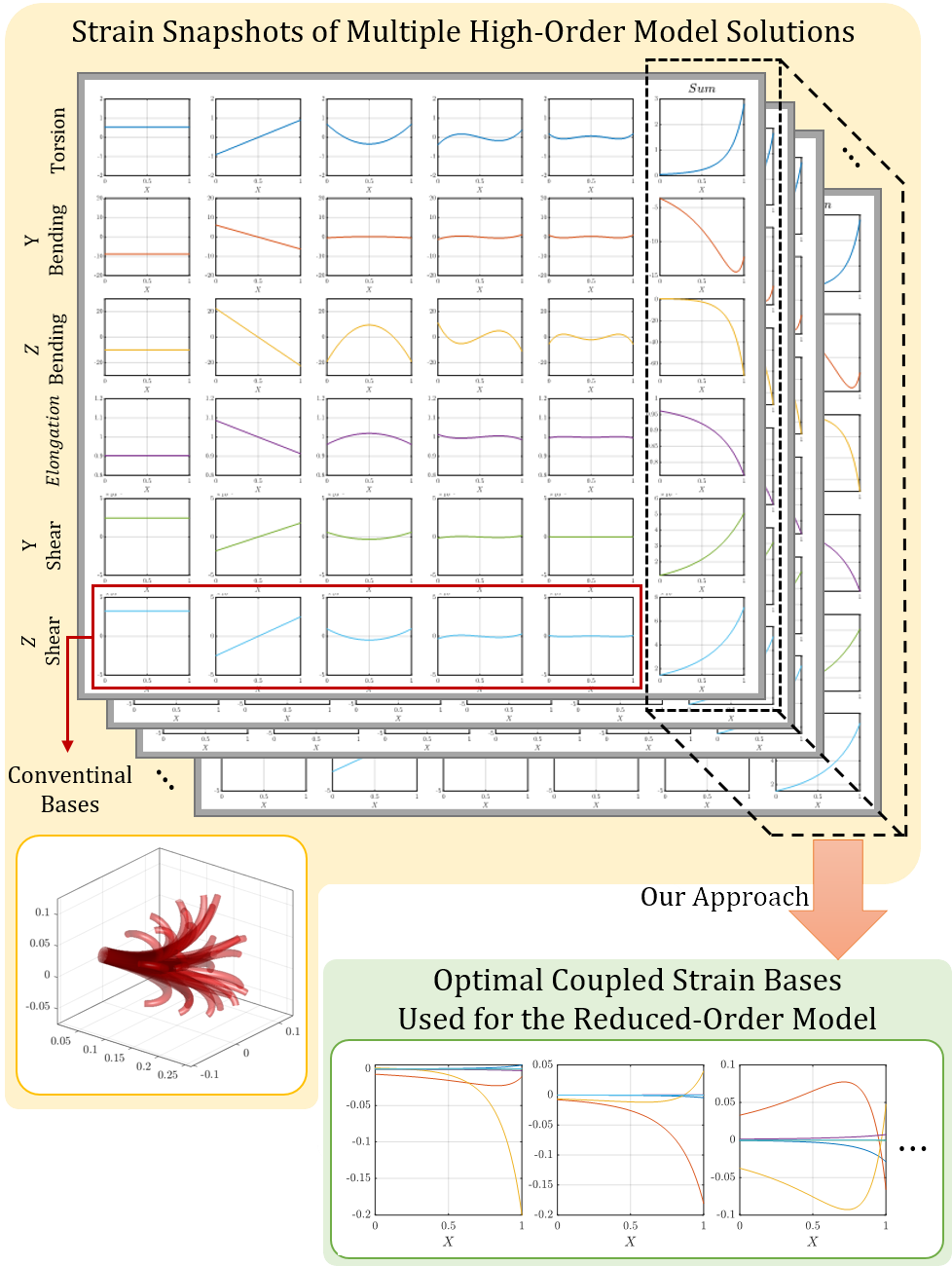}
    \caption{\small An overview of our order reduction method. The last element of each row in the snapshots, which is the strain field solution, is the sum of the preceding elements, each governed by separate coordinates. This example shows the reduction of a single soft, slender manipulator with multiple actuators. }
    \label{fig::OpeningFig}
\end{figure}

\IEEEPARstart{T}{HE} interest in soft robotics emerges from the pressing need to create robotic systems capable of seamlessly adapting to complex and dynamic environments \cite{laschi2016soft}. Soft robots, characterized by their compliant and flexible nature, offer a remarkable solution to overcome the constraints of rigid robots, which are limited by a finite number of degrees of freedom. Soft robots can deform and change their shape, allowing them to adapt to a wide range of environments, handle fragile objects without causing damage, interact directly with humans, and engage in collaborative tasks in shared workspaces, among other capabilities. The importance of these robots has attracted the interest of a great number of researchers in recent decades, especially in
the development of mathematical models \cite{armanini2023soft} that can accurately describe their behavior without requiring heavy computational loads.

%



Slender soft robots, inspired by animal tentacles, trunks, and slender biological organisms such as snakes and certain types of plants, constitute a common class of soft robots. Various approaches have been explored to model these systems. Finite-Element Methods (FEM) are widely used to model this type of soft robot \cite{xavier2021finite}, \cite{tawk2020finite}. However, their computational cost is very high, and solvers may struggle to converge on a solution. To reduce the computational load in simulating these systems, many researchers in the field of soft robotics have developed mathematical models that consider one physical dimension longer than the other two, such as in tentacle-like systems. These models effectively approximate the entire configuration by neglecting volumetric deformations and focusing on the behavior of the central axis. The exact description of the dynamics of rod-like structures can be obtained by using the continuous Kirchhoff-Clebsch-Love and Cosserat rod theories \cite{tummers2023cosserat}. Recent works have proposed simplifying these rod-like models by restricting the range of possible shapes to a finite-dimensional functional space. Among these models is the Piecewise Constant Curvature (PCC) model, which is the most commonly used in the field of soft robotics \cite{godage2016dynamics},\cite{webster2010design}. The PCC model describes the central axis of the soft robot as a finite number of stitched circular arcs. \cite{renda2018discrete} introduces the Piecewise Constant Strain (PCS) model, which extends PCC models to encompass torsion and shears, in addition to curvature and elongation and generalize circular arcs to screw arcs. In \cite{li_PLS_2022}, the Piecewise Linear Strain (PLS) model was introduced utilizing linear strain bases in a fashion similar to most FEM approaches. Furthermore, the Geometric Variable Strain (GVS) model \cite{renda2020geometric}, \cite{boyer2020dynamics} was introduced. This approach proposes that the soft manipulator can be completely described by a finite set of strain bases, which are not necessarily constant. In such models, the coefficients of these bases play the same role as that of the joint vector or the Degrees of Freedom (DOFs) in traditional robotics. The aforementioned strain-based modeling approaches offer several advantages over other models. They provide excellent accuracy even for extreme deformations by treating the geometry and balance equations exactly. Despite this accuracy, they maintain low computational complexity compared to FEM models. Moreover, these models are generic and do not rely on specific assumptions about the robot's cross-section or actuator path, making them applicable to a wide range of designs. Importantly, unlike position-based models, these models can accommodate discontinuities in the strain fields, providing more flexibility in their parameterization.



Although smaller in dimension compared to other approaches such as FEM, previously discussed models usually result in large, high-dimensional models that are still expensive to solve. This led to an interest in improving the computational efficiency of simulating soft robots, especially for real-time control applications, introducing the field of Model Order Reduction (MOR). These methods are widely used in the computational mechanics community to reduce the computational cost of FEM simulations. They enable the reduction of the degrees of freedom in the computed mechanical system while maintaining precision. One of the main methods for accomplishing this is Proper Orthogonal Decomposition (POD) \cite{kerschen2005method}. This method, based on the collection of system configuration data, simplifies a complex set of interconnected variables by transforming them into a reduced set of uncorrelated variables, capturing the majority of the variability in the original dataset. Some previous works that use the MOR technique for FEM simulations can be found in \cite{chenevier2018reduced} for simple soft robots or \cite{goury2018fast,goury_real-time_2021} for more generic and complex soft robots, which was implemented for a control application in \cite{thieffry_dynamic_nodate}. Another interesting work that uses FEM with model order reduction is \cite{tonkens2021soft}, which proposes an approach for constrained optimal control of soft robots. An equivalent approach used for dimensionality reduction and feature extraction in data analysis is principal component analysis. It has been effectively applied to find hand synergies by using a low-dimensional subspace of the hand DOFs space (see, e.g., \cite{santello1998postural}, \cite{ciocarlie2009hand}). Another popular MOR technique is Balanced Model Reduction (BMR), which takes into account the input matrix to obtain the most controllable reduced modes \cite{yoon_passive_2019}. The authors propose a fast simulation framework for soft objects with intermittent contacts, performing BMR for each contact mode and switching between the reduced-order models based on the contact areas/locations. 

Many researchers nowadays focus on data-driven modeling and control of soft robots to overcome the challenges associated with their inherent nonlinearities and high DOFs. Koopman operator theory has emerged as a promising approach for learning globally linear representations of nonlinear dynamics, enabling the application of linear control techniques to soft robotic systems \cite{bruder_data-driven_2021}. This approach has been successfully applied to the control of soft continuum arms, demonstrating improved performance compared to traditional model-based controllers \cite{wang_improved_2023,haggerty_control_2023}. Additionally, spectral analysis of the learned Koopman operators has provided insights into the physical characteristics of the system, such as oscillation frequencies and energy distributions \cite{komeno_deep_2022}.
Other data-driven approaches, such as machine learning and neural networks, have been used to model soft robots \cite{kim2021review}. Mainly, recurrent neural networks (RNNs) have been used to model the complex dynamics of soft robotic manipulators, as demonstrated in several recent works \cite{thuruthel_learning_2017,thomas_george_thuruthel_first-order_2020,tariverdi_recurrent_2021}. These learned models have then been employed for developing open-loop and closed-loop controllers for soft manipulators \cite{thuruthel_stable_2018,thuruthel_model-based_2019}. Spectral Submanifold Reduction (SSMs) was presented in \cite{alora_data-driven_2023} as a data-driven approach for learning low-dimensional models of high-dimensional robots on spectral submanifolds by learning the dynamics of generic low-dimensional attractors. Unlike analytical models, such data-driven models are completely learned from data and are representative of the specific datasets used for their training, making them not generalizable nor interpretable.  


Combining data-driven approaches with physics models can lead to more streamlined, computationally efficient, and system-specific mathematical models. A notable trend in the field of data-driven modeling is the integration of data-driven tools to enhance traditional physics-based models. Unlike approaches that rely entirely on data to model a system or those that incorporate the known physics of a phenomenon to guide the training of data-driven models \cite{raissi_physics-informed_2019,karniadakis_physics-informed_2021}, data-enhanced physics models use data-driven tools to improve accuracy, reduce computational complexity, and/or lower the system's order \cite{lepri_neural_2023}. We refer to systems that integrate both physical principles and data-driven elements as \emph{hybrid} systems. While physics models offer the best interpretability and generalization, they are usually costly to solve, a disadvantage that data-driven models are able to solve. Nevertheless, data-driven models require a lot of prior work and lack generalization. Hybrid models promise the best of both worlds. A qualitative comparison between physics-based, hybrid, and fully data-driven models is provided in Table \ref{tab::Comparison}. 


\begin{table}[]
\centering

\caption{\small Qualitative comparison between physics, hybrid and data-driven models. Plus signs mean positive traits, while dashes mean negative traits.}
\begin{tabular}{ l  c  c  c}
\hline
 & Physics & Hybrid & Data-Driven \\ \hline
Computational Cost & - - & + & ++ \\ \hline
DOFs & - & ++ & + \\ \hline
Accuracy  & + & + & ++ \\ \hline
Generalization & ++ & + & - \\ \hline
Offline Work Demand & ++ & + & - - \\ \hline
\end{tabular}
\label{tab::Comparison}
\end{table}

In this paper, we unite two primary principles developed thus far: the aforementioned strain-based GVS model for the mechanics of soft robots and the POD method to find the optimal strain bases required to describe the system. Despite being the state-of-the-art general purpose soft robot model, the GVS still exhibits high dimensionality and computational load. By combining it with POD, we achieve significantly reduced-order, interpretable, and generalizable models. Our approach differs from other POD-based ROMs in that we do not need to compute anything related to the Higher-Order Model then project it to the reduced subspace; instead, we obtain coupled modes or synergies capable of efficiently describing the system using the fewest possible DOFs and construct the ROM directly using any principle of mechanics.
Essentially, our approach can be viewed as a way to reduce the dimensions of the configuration space with minimal accuracy deterioration. In addition, it is hybrid in the sense that it integrates both physics-based modeling and data-driven techniques. It is crucial to emphasize that, in contrast to our POD approach, other model order reduction techniques use the data to reduce the equations of motion directly, often assuming a specific model structure.

The significance of incorporating POD with the GVS model can be summarized in three main points: 1) Coupling the six strains into a single mode analytically is feasible for simple configurations but challenging for complex ones. Our method enables the acquisition of these coupled interpretable modes for various shapes, actuator paths, and loading conditions. 2) Beyond identifying coupled strain bases, POD optimizes the system description to capture most of the behavior with the fewest DOFs, transitioning from generic bases (usually high-order polynomials) to optimal bases. 3) The GVS framework extends traditional geometric rigid robotics models, unifying the treatment of soft body strains and rigid joint twists. This allows for direct application of the reduction technique across soft, rigid, and hybrid systems, with our work being the first to find the synergies of soft and hybrid systems.

By leveraging the strengths of both the GVS model and the POD method, we create a highly efficient and accurate representation of the robot's mechanics. We thoroughly analyze the performance of the proposed ROM in terms of accuracy and computational demand, showing the capability of achieving real-time simulations in specific cases. Additionally, we apply the proposed reduction approach to a variety of challenging robotic prototypes, including cable-driven soft, hyper-redundant rigid, and hybrid soft-rigid prototypes, to showcase its versatility. Additionally, we implement a shape estimation algorithm using a few position markers, enabled by the knowledge and insights acquired from the system reduction, demonstrating impressive qualitative and quantitative performance. The shape estimation algorithm is tested on a six-actuator, multi-section, soft robotic manipulator, further validating the effectiveness and applicability of our approach in real-world applications. Finally, we experimentally validate the ROM on a parallel hybrid soft-rigid system, showcasing it's ability to represent the HOM and the real system with minimal accuracy deterioration.


The structure of this paper is organized as follows: Section II provides an overview of the GVS approach. Section III introduces the POD method and explains how we integrate it with the GVS to create optimal reduced-order models. In Section IV, we present the implementation and analysis of our reduction method applied to various soft manipulators. Section V extends the application of our method to hyper-redundant rigid and hybrid prototypes. In Section VI, we showcase the experimental application of the ROM to a soft manipulator shape estimation problem, in addition to ROM validation experiments on a hybrid soft-rigid system. Finally, Section VII concludes the paper by summarizing our work and discussing potential future research directions.


\section{Geometric Variable Strain Model}

In this section, we recall the essential components of the Geometric Variables Strain (GVS) approach used for simulating a single soft manipulator. Additionally, we provide a summary of its extension to include a hybrid multibody system. For a detailed description of the GVS approach, the reader is referred to \cite{Mathew_SoRoSim2021,Anup_IJRR}. 


\subsection{Kinematics}\label{sec::Kinematics}
\begin{figure*}
    \centering
    \includegraphics[width=0.75\linewidth]{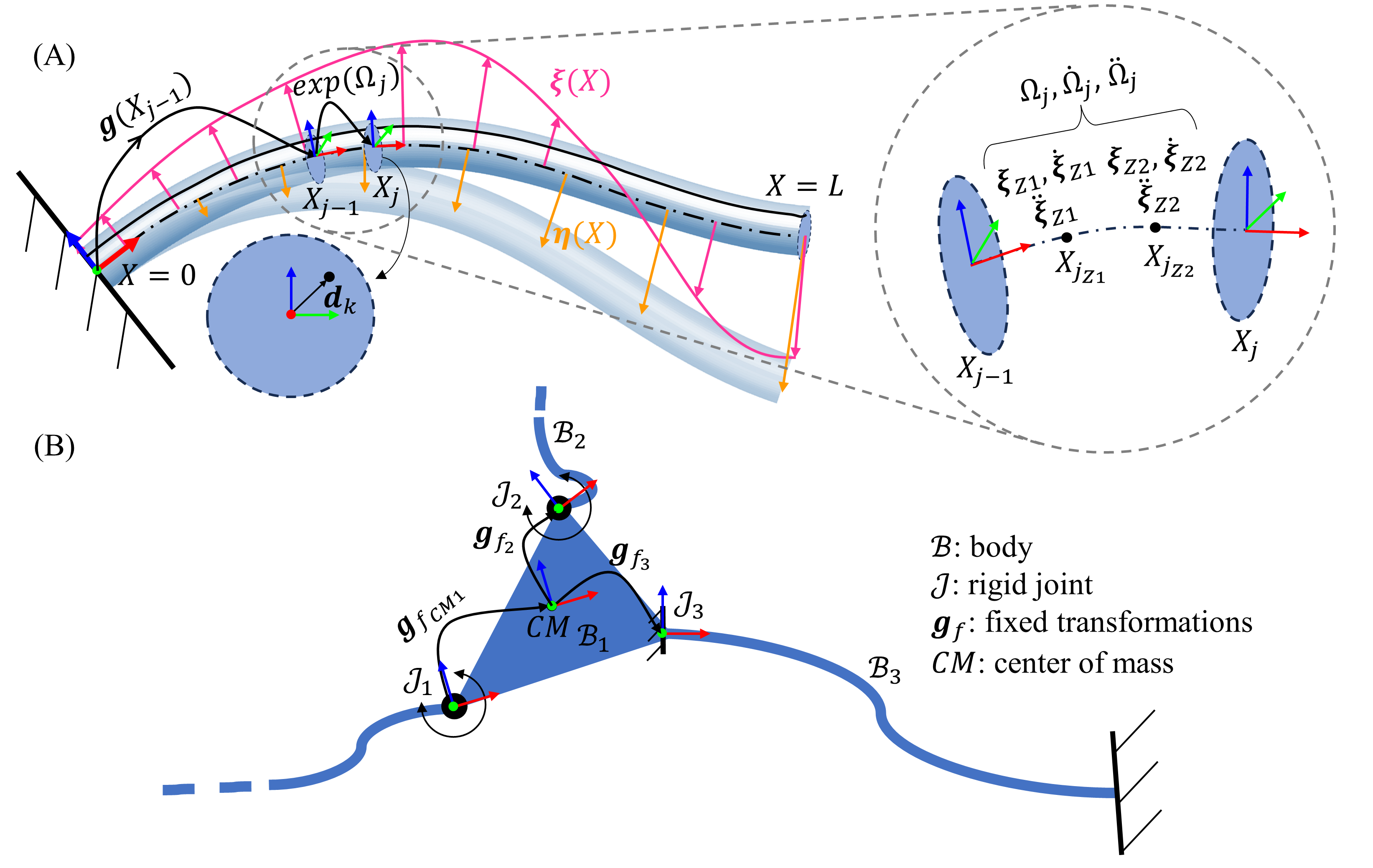}
    \caption{\small Schematics of the GVS model: (A) Soft manipulator with a representative strain and velocity field. The recursive computation from $X_{j-1}$ to $X_j$ using two Zannah points, $Z1$ and $Z2$, is highlighted. (B) A generic hybrid multibody system with branched and close-loop joints.}
    \label{fig::model}
\end{figure*}
A soft manipulator can be modeled as a Cosserat rod, a continuous stack of non-deformable cross-sections parameterized by a curvilinear abscissa \(X \in [0,L]\) where \(L\) is the length of the rod (Fig. \ref{fig::model}A). By rigidly attaching local coordinate frames to each of these cross-sections, with the local x-axis perpendicular to the cross-section, the configuration space can be fully defined as a directed spatial curve, $\boldsymbol{g}(\bullet):X\rightarrow \boldsymbol{g}(X) \in SE(3)$. The homogenous transformation matrix $\boldsymbol{g}(X)$ is given by:
\begin{equation}
    \boldsymbol{g}(X) = \left[ \begin{array}{cc} \boldsymbol{R} & \boldsymbol{r} \\ \boldsymbol{0}^{1\times 3} & 1 \end{array} \right] 
    \label{eqn::homogenousTM_g}
\end{equation}
\noindent where \(\boldsymbol{r}(X) \in \mathbb{R}^3\) is the position vector to the origin of the local frame and the rotation matrix \(\boldsymbol{R}(X) \in SO(3)\) represents the orientation of the local frame with respect to the spatial frame. The spatial $(\cdot)'$ and temporal $\dot{(\cdot)}$ partial derivative of Equation \eqref{eqn::homogenousTM_g} defines the screw strain ($\boldsymbol{\xi}$) and screw velocity ($\boldsymbol{\eta}$) of the Cosserat rod:
\begin{equation}\label{eqn::gprime} \boldsymbol{g}'(X)=\boldsymbol{g}\hat{\boldsymbol{\xi}}; \hspace{0.7cm} \dot{\boldsymbol{g}}(X)=\boldsymbol{g}\hat{\boldsymbol{\eta}}
\end{equation}

\noindent where the operator \(\hat{(\bullet)}\) indicates the isomorphism from $\mathbb{R}^6$ to $se(3)$. \(\hat{\boldsymbol{\xi}}(X)\). Using reference strain or the stress-free configuration derivative, $\boldsymbol{\xi}^*(X)$, we define,
\begin{equation*}
    \hat{\boldsymbol{\xi}} - \hat{\boldsymbol{\xi}}^* = \left[ \begin{array}{cc} \tilde{\boldsymbol{k}} & \boldsymbol{l} \\ \boldsymbol{0}^{1\times 3} & 0 \end{array} \right] \in se(3), \; 
\end{equation*}
\noindent
where $\boldsymbol{k}(X) = [k_x,\; k_y,\; k_z]^T$ and $ \boldsymbol{l}(X) = [l_x,\; l_y,\; l_z]^T$ are the angular and linear strains in the local frame. The operator \(\tilde{(\bullet)}\) indicates the isomorphism between \(\mathbb{R}^3\)
and the algebra of skew-symmetric matrices \(so(3)\). The velocity and strain twists are related through the equality of the mixed partial derivatives, yielding:
\begin{equation}\label{eqn::vel}   \boldsymbol{\eta}'=\dot{\boldsymbol{\xi}}-\text{ad}_{\boldsymbol{\xi}}\boldsymbol{\eta}
\end{equation}
\noindent where \(\text{ad}_{(\boldsymbol{\bullet})}\) is the adjoint operator of \(se(3)\). 
Integrating equations \eqref{eqn::gprime} (the first part) and \eqref{eqn::vel} over space yields:

\begin{equation}\label{eq::int_kinem_pos}
\bm{g}(X) = \text{exp} \left( \widehat{\bm{\Omega}}(X) \right) \; \text{,}
\end{equation}
\begin{equation}\label{eq::int_kinem_eta}
    \boldsymbol{\eta}(X) =  \text{Ad}_{\boldsymbol{g}^{-1}}\int_0^{X} \text{Ad}_{\boldsymbol{g}} \dot{\boldsymbol{\xi}} ds
\end{equation}
where \(\text{Ad}_{\boldsymbol{g}}\) is the adjoint representation of the homogenous matrix \(\boldsymbol{g}\), and $\bm{\Omega}$ is the Magnus expansion of $\bm{\xi}(X)$. The GVS model discretizes the Cosserat rod strains to introduce the generalized coordinates. The continuous strain field of the rod is modeled using a finite set of strain bases:
\begin{equation}
    \boldsymbol{\xi}(X) - \boldsymbol{\xi}^*(X) = \boldsymbol{\Phi}_\xi(X)\boldsymbol{q} 
    \label{eqn::discretizedStrain}
\end{equation}
where \(\boldsymbol{\Phi}_\xi(X)\in \mathbb{R}^{6 \times n}\) is the matrix function whose columns form the basis for the strain field, \(\boldsymbol{q}\in \mathbb{R}^n\) (with \(n\) being the number of generalized coordinates) is the vector of generalized coordinates which spans the chosen basis. Substituting Equation (\ref{eqn::discretizedStrain}) into equation (\ref{eqn::vel}) we get,
\begin{equation}
\label{eq::Jacobian}
    \boldsymbol{\eta}(X) = \mathrm{Ad}_{\bm{g}(X)}^{-1} \int_0^X \mathrm{Ad}_{\bm{g}} \bm{\Phi}_{\xi} ds \dot{\bm{q}} = \boldsymbol{J}(\boldsymbol{q},X)\dot{\boldsymbol{q}}
\end{equation}
where \(\boldsymbol{J}\in \mathbb{R}^{6 \times n}\) is the geometric Jacobian. A time derivative of \eqref{eq::Jacobian} defines the acceleration twist $\dot{\boldsymbol{\eta}}$ and $\dot{\boldsymbol{J}}$:
\begin{equation}
\label{eq::accleration}
    \dot{\boldsymbol{\eta}}(X) = \boldsymbol{J}(\boldsymbol{q},X)\ddot{\boldsymbol{q}}+\dot{\boldsymbol{J}}(\boldsymbol{q},\dot{\bm{q}},X)\dot{\boldsymbol{q}}
\end{equation}
For a general $\bm{\xi}(X)$, the configuration, Jacobian and its derivative cannot be computed explicitly. Instead, these elements are determined through a recursive formulation that is applied from one discrete point to the next, employing the Zannah quadrature approximation of the Magnus expansion. The procedure involves the computation of strain and its time derivatives at Zannah quadrature points that lie in between the discrete points (Fig. \ref{fig::model}A). The details of the recursive scheme can be found in \cite{Anup_IJRR,GeomExct_Renda2022}. 
\subsection{Dynamics and Statics}\label{sec::DynStat}
Projecting the free dynamics of the Cosserat rod \cite{boyer2020dynamics} using the geometric Jacobian through D'Alembert's principle and integrating over the length of the soft link yields the generalized dynamics of the system in the standard Lagrangian form \cite{AnupSoRoSim}:
\begin{equation}
\label{eq::gendynamics}
\bm{M}(\bm{q})\ddot{\bm{q}} + (\bm{C}(\bm{q},\dot{\bm{q}}) + \bm{D}(\bm{q}))\dot{\bm{q}} + \bm{K}\bm{q} = \bm{B}(\bm{q})\bm{T} + \bm{F}(\bm{q}) \; \text{,}
\end{equation}
where $\bm{M}(\bm{q})$, is the generalized mass matrix, $ \bm{C}(\bm{q}, \dot{\bm{q}})$ is the Coriolis matrix, $\bm{D}(\bm{q})$ is the elastic damping matrix, $ \bm{K}$ is the stiffness matrix, $ \bm{B}(\bm{q})$ is the actuation matrix, $ \bm{F}(\bm{q}) $ is the vector of generalized external forces, and $\bm{T}$ is the vector of applied actuation strengths. 
To compute $\bm{K}$ and $\bm{D}$, a linear Hooke-like elastic and Kelvin-Voigt damping model is used \cite{renda2014dynamic}. While this damping model does not account for hysteresis or creep, it fits our systems as such effects are present in negligible magnitudes. For soft bodies that are internally actuated by threadlike actuators (Fig. \ref{fig::model}A), the internal actuation wrench exerted by the actuator on the rod's centerline is modeled as:
\begin{equation}\label{ref::cable}
    \boldsymbol{\mathcal{F}}_a(X) = \sum_{k=1}^{n_a} \left[\begin{array}{c}
         \tilde{\boldsymbol{d}}_k\boldsymbol{t}_k  \\
         \boldsymbol{t}_k
    \end{array} \right] T_k = \boldsymbol{\Phi}_a(\boldsymbol{q},X)\boldsymbol{T}
\end{equation}
where \(\boldsymbol{\Phi}_a(\boldsymbol{q},X)\in \mathbb{R}^{6\times n_a}\) is the actuation basis matrix, \(n_a\) is the number of actuators, \(\boldsymbol{d}_k(X) = [0,\;p_{y_k},\;p_{z_k}]^T\) is the local cross-sectional position of actuator \(k\), and \(\boldsymbol{t}_k(X)\) is the unit vector tangent to the actuator's path. Fig.\ref{fig::model}A shows a generic actuator path \(\boldsymbol{d}_k(X)\) visually as the solid black line.

The generalized static equilibrium equation is derived by equating all time derivatives ($\dot{\bm{q}}$, $\ddot{\bm{q}}$) in \eqref{eq::gendynamics} to zero:
\begin{equation}
\label{eqn::genStcEqn}
     \boldsymbol{K}\boldsymbol{q} = \boldsymbol{B}(\boldsymbol{q})\boldsymbol{T} + \boldsymbol{F}(\boldsymbol{q})
\end{equation}
\subsection{Hybrid Multibody System}
In the GVS framework, the distributed strain of the soft body and the twist of a rigid joint are analyzed through a unified approach. This method facilitates the modeling of hybrid multibody systems by integrating rigid and soft links. The formulation also supports the modeling of robots that are branched chain structures or closed-chain systems (Fig. \ref{fig::model}B). The form of generalized dynamics \eqref{eq::gendynamics} and statics \eqref{eqn::genStcEqn} remain the same. However, the components are calculated by summing the contributions from all links, each projected using corresponding Jacobians. Readers may refer to Appendix \ref{app:B} for their formulae.

\par

For closed-chain systems, the methodology outlined in \cite{Armanini_TRO2021} can be adopted to include closed-loop constraints. This involves integrating constraint forces into the generalized dynamics and adding constraint equations in Pfaffian form (at the velocity level). In the case of statics, these constraints are represented at the position level instead.
\par

The generalized dynamics is solved using an ODE integrator, while the static equilibrium is obtained by numerical root-finding methods. To implement and solve the GVS model, we used the MATLAB toolbox SoRoSim \cite{Mathew_SoRoSim2021}. The toolbox allows the creation of soft and rigid links with rigid joints and the assembly of these links to form serial, branched, or closed-chain robots. It uses MATLAB functions, such as $ode45$ for the time integration and $fsolve$ for solving static equilibrium. Readers interested in the validation of the GVS model for static and dynamic scenarios are referred to \cite{Mathew_SoRoSim2021}.

\section{Reduction of the GVS using POD}\label{sec::POD}

Our model order reduction technique is based on the snapshot Proper Orthogonal Decomposition (POD), a well-established method for identifying optimal basis sets in ROMs. In the POD framework, we first collect high-fidelity data snapshots that capture the spatio-temporal characteristics of the quantity of interest. These snapshots are then vectorized and assembled into what is termed the \emph{snapshot matrix} \( \boldsymbol{\Xi} \in \mathbb{R}^{p \times s} \), where $p$ is the number of sampling points, and $s$ is the number of snapshots. The snapshot matrix \( \boldsymbol{\Xi} \) is then subjected to Singular Value Decomposition (SVD), given by:
\begin{equation}
\label{eq::SVD} 
\boldsymbol{\Xi} = \boldsymbol{U \Sigma V^T}
\end{equation}
where \( \boldsymbol{U} \in \mathbb{R}^{p \times p} \) and \( \boldsymbol{V} \in \mathbb{R}^{s \times s} \) are orthogonal matrices that contain the left and right singular vectors, respectively, and the rectangular diagonal matrix \( \boldsymbol{\Sigma} \in \mathbb{R}^{p \times s} \) comprises the singular values of \( \boldsymbol{\Xi} \). The columns of \( \boldsymbol{U} \) are employed as the optimal basis vectors for the ROM and are ordered in accordance with their associated singular values in descending order, and we will address them as the decomposition modes. Essentially, the magnitude of a singular value serves as an indicator of the extent to which its corresponding mode encapsulates information from the original data. While it is possible to reconstruct lower-rank approximations of \( \boldsymbol{\Xi} \) by utilizing only the leading \( r \) columns of \( \boldsymbol{U} \) and \( \boldsymbol{V} \), as well as the principal \( r \times s \) submatrix of \( \boldsymbol{\Sigma} \) in equation (\ref{eq::SVD}), our focus is not on data dimensionality reduction. Instead, we are particularly interested in the identification of optimal bases that effectively represent the original high-fidelity data.
\subsection{Reduction in Continuous Domain}
Incorporating the POD methodology into the GVS model is a relatively straightforward process. The objective is to identify an optimal basis matrix \( \boldsymbol{\Phi}_{\xi_{\mathcal{O}}} \) that efficiently captures the system behaviour using a minimal number of generalized coordinates \( \boldsymbol{q} \). This approach condenses high-order descriptions into a few key components. Moreover, in our earlier studies, we manually specified the basis matrix based on a chosen order. However, in that approach, each column of the basis matrix contained only a single non-zero element to ensure the independence of multiple strain modalities (e.g., bending and elongation) on the same coordinate, as identifying a priori coupling between these modes is a challenging task. To illustrate this, let \( \mathcal{L}_\mathfrak{i}(X) \) represent a Legendre polynomial of order \( \mathfrak{i} \), which is scaled to the interval \([0, L]\) rather than the conventional interval \([-1, 1]\). A basis matrix can then be defined as follows:
\begin{equation}\label{eq::globalbasis_Leg}\small\boldsymbol{\Phi}_{\xi}=\left[ \begin{array}{ccccc} \mathcal{L}_0(X) & \mathcal{L}_1(X)  & 0 & 0 & 0 \\ 0 & 0  & 0 & 0 & 0 \\ 0 & 0  & \mathcal{L}_0(X) & \mathcal{L}_1(X) & \mathcal{L}_2(X)\\ 0 & 0  & 0 & 0 & 0\\ 0 & 0  & 0 & 0 & 0\\ 0 & 0  & 0 & 0 & 0\end{array} \right]
\end{equation}
This matrix specifies that the torsion is represented by the first two Legendre polynomials, while bending about the z-axis is represented by the first three Legendre polynomials. The remaining strain fields are assumed to be negligible. It is apparent that each column is responsible for a single strain modality. Our proposed approach overcomes this limitation, while identifying a few primary contributing modes, thereby significantly reducing the number of generalized coordinates.

To start, let \( \boldsymbol{k}_x \) represent a discrete form of the continuous torsional strain field \( k_x(X) \) as follows:
\begin{equation}\small
\boldsymbol{k}_x := [k_x(0) , k_x(X_1) , k_x(X_2) , \dots , k_x(X_{p-2}) , k_x(L)]^T \in \mathbb{R}^{p},
\end{equation}
where, \( X_1,X_2,\ldots,X_{p-2} \) denote intermediate points that are sampled arbitrarily along the abscissia \( X \). Applying the same sampling technique to the other strain modalities yields corresponding column vectors \( \boldsymbol{k}_y \), \( \boldsymbol{k}_z \), \( \boldsymbol{l}_x \), \( \boldsymbol{l}_y \), and \( \boldsymbol{l}_z \) \(\in\) \( \mathbb{R}^p \). 

The different strain modalities are vertically concatenated to form a single column vector of length \( 6p \), which constitutes a snapshot of the system in the following manner:
\begin{equation}
    \Xi = [\boldsymbol{k}_x^T, \boldsymbol{k}_y^T,  \boldsymbol{k}_z^T,  \boldsymbol{l}_x^T,  \boldsymbol{l}_y^T, \boldsymbol{l}_z^T]^T \in \mathbb{R}^{6p}.
\end{equation}
Then the snapshot matrix is assembled as:
\begin{equation}
    \boldsymbol{\Xi} = [\Xi_1,\Xi_2,\dots,\Xi_s] \in \mathbb{R}^{6p\times s}.
\end{equation}
where $s$ is the total number of snapshots. The reason behind the vertical concatenation can be explained by the analogy between Equations (\ref{eqn::discretizedStrain}) and (\ref{eq::SVD}). It can be seen that \( \boldsymbol{\Phi}_\xi \) is analogous to \( \boldsymbol{U} \), and \( \boldsymbol{q} \) is analogous to \( \boldsymbol{\Sigma V^T} \). In this context, the columns of \( \boldsymbol{V} \) serve as scaling factors for the corresponding modes (columns of \( \boldsymbol{U} \)). These scaling factors describe the evolution of each mode in every snapshot, a role analogous to that of \( \boldsymbol{q} \) in relation to \( \boldsymbol{\Phi}_\xi \). Therefore, vertically concatenating the sampled strain vectors results in coupled modes that are governed by the same generalized coordinate.

Upon completing the SVD of the snapshot matrix \( \boldsymbol{\Xi} \), the resulting \( \boldsymbol{U} \) matrix is then:
\begin{equation}\label{eq::U_SVD}
    \boldsymbol{U} = [\boldsymbol{u}_1 , \boldsymbol{u}_2 , \dots , \boldsymbol{u}_{6p}] \in \mathbb{R}^{6p\times 6p}
\end{equation}
and the $c^{th}$ mode is composed of:
\begin{equation}
    \boldsymbol{u}_c = \left[{\boldsymbol{u}^{k_x}_c}^T , {\boldsymbol{u}^{k_y}_c}^T , {\boldsymbol{u}^{k_z}_c}^T , {\boldsymbol{u}^{l_x}_c}^T , {\boldsymbol{u}^{l_y}_c}^T , {\boldsymbol{u}^{l_z}_c}^T \right]^T \in \mathbb{R}^{6p}
\end{equation}
This yields coupled optimal basis vectors in discrete form, which can subsequently be transformed into continuous fields $(u^{k_x}_c(X), u^{k_y}_c(X), u^{k_z}_c(X), u^{l_x}_c(X), u^{l_y}_c(X), u^{l_z}_c(X))$ through interpolation for inclusion in equation (\ref{eqn::discretizedStrain}). The optimal basis matrix can then be constructed for $n$ generalized coordinates by selecting $n$ columns of (\ref{eq::U_SVD}) and rearranging them as: 
\begin{equation}\label{eq::PODbasis}
\boldsymbol{\Phi}_{\xi_\mathcal{O}}(X)=\left[ \begin{array}{cccc} u^{k_x}_1(X) & u^{k_x}_2(X)  & \dots & u^{k_x}_n(X) \\ u^{k_y}_1(X) & u^{k_y}_2(X)  & \dots & u^{k_y}_n(X) \\ u^{k_z}_1(X) & u^{k_z}_2(X)  & \dots & u^{k_z}_n(X) \\ u^{l_x}_1(X) & u^{l_x}_2(X)  & \dots & u^{l_x}_n(X) \\ u^{l_y}_1(X) & u^{l_y}_2(X)  & \dots & u^{l_y}_n(X) \\ u^{l_z}_1(X) & u^{l_z}_2(X)  & \dots & u^{l_z}_n(X) \end{array} \right]
\end{equation}
It is worth noting that now -as expected- a single coordinate is capable of producing all modalities of strain, i.e. coupling is achieved. This concept is the soft variant of synergies in rigid robots \cite{santello_hand_2016}, but extended for spatial coupling and multi-directional strains. It is also worth mentioning that the SVD is computationally very cheap, having the decomposition done in a matter of seconds for large snapshot matrices. For example, a snapshot matrix dealt with in this paper, as discussed in Section \ref{sec::6Act}, with 6k snapshots and an abscissa discretization of 102 points, i.e. \( \boldsymbol{\Xi} \in \mathbb{R}^{612\times 6000} \), was decomposed in just 1.2 s when run on an average computer.
\subsection{Reduction in Discrete Domain}\label{sec::POD_Hybrid}
To integrate the reduction approach with the discrete domain implementation of the kinematics discussed at the end of Section \ref{sec::Kinematics}, we follow a simpler process. This process facilitates the application to rigid joints and hybrid soft-rigid multibody systems. Assuming that there are $p$ points at which Cosserat rod strains of all soft bodies and joint twists of all rigid joints are computed, a concatenated vector of strains and twists can be defined as:
\begin{equation}
\bar{\bar{\bm{\xi}}} := [\bm{\xi}_1^T-\bm{\xi}_1^{*T},\bm{\xi}_2^T-\bm{\xi}_2^{*T},...,\bm{\xi}_p^T-\bm{\xi}_p^{*T}]^T \in \mathbb{R}^{6p}
\end{equation}
where $\bm{\xi}_j-\bm{\xi}_j^*$ is the strain at the $j^{th}$ computational point. Note that the computational points include the rigid joints, the quadrature points of the space integration, and the Zannah quadrature points used for the approximation of Magnus expansion.

Applying POD and reduction of multiple snapshots of $\bar{\bar{\bm{\xi}}}$ gives an optimal concatenated basis ${\boldsymbol{\Phi}_{\bar{\bar{\xi}}_{\mathcal{O}}}} \in \mathbb{R}^{6p\times n}$. The concatenated vector of reduced order strain is then given by:

\begin{equation}
\label{eq:reducedstrain}
\bar{\bar{\bm{\xi}}} = {\boldsymbol{\Phi}_{\bar{\bar{\xi}}_{\mathcal{O}}}}\bm{q}+\bar{\bar{\bm{\xi}}}^*
\end{equation}
where $\bm{\xi}_C^*$ is the concatenated reference strain, $\bm{q} \in \mathbb{R}^{n}$ is the DOF of the reduced order model. The strain value at the $j^{th}$ computational point can be extracted from \eqref{eq:reducedstrain} according to:
\begin{equation}
\bm{\xi}_j = \bm{\mathcal{E}}_j{\boldsymbol{\Phi}_{\bar{\bar{\xi}}_{\mathcal{O}}}}\bm{q}+\bm{\xi}_j^*
\end{equation}
where, $\bm{\mathcal{E}}_j = [\bm{0}^{6\times 6(j-1)}\;\bm{I}_6\;\bm{0}^{6\times 6(p-j)}]$. The reduced basis for the $j^{th}$ computational point is given by $\bm{\mathcal{E}}_j{\boldsymbol{\Phi}_{\bar{\bar{\xi}}_{\mathcal{O}}}}$. Since the computational points include all quadrature points, this process avoids the interpolation step during space integration. The formulae for coefficients of generalized statics and dynamics remain the same, except for $\bm{D}$, $\bm{K}$, and $\bm{B}$. The updated formulas for the reduced basis are provided in Appendix \ref{app:B}.

\subsection{Truncation Evaluation}\label{sec::truncation}

A standard approach to evaluate the relative energy content of each mode in POD involves comparing the \(c^{th}\) singular value \( \sigma_c \) to the remaining singular values. A commonly employed metric for this assessment is given by the equation:
\begin{equation}
\varepsilon(r) = \frac{\sum_{c=1}^{r} \sigma_c^2}{\sum_{c=1}^{s} \sigma_c^2}.
\end{equation}
Here \( \varepsilon \) quantifies the proportion of total relative energy captured when using the first \( r \) modes of the decomposition. It's crucial to keep in mind that while this equation offers insights into the accuracy of the rank \( r \) reconstruction of the original data and the relative importance of each mode, for base selection in model reduction, configuration-based metrics are required instead. This is due to the differences in scale between various strains. For instance, elongation strains are typically one or two orders of magnitude smaller than bending strains. Therefore, when decomposing all six strains simultaneously, less importance is given to elongation strains, which might have lower relative energy and appear dominantly in a later mode. This leads to the assumption that \(\varepsilon(r)\) captures most of the variance, while overlooking a crucial elongation mode that could significantly alter the behavior of the reduced model.

\subsection{Data Generation}
To generate the data necessary for our model order reduction method, we employ different strategies depending on the robotic system and simulation scenario:
\begin{enumerate}
    \item Static Simulations: For static cases, we sample the actuation space comprehensively. This involves sampling each dimension in the actuation space, creating a hypercubic point cloud, then including all combinations to cover all possible configurations of the robot.
    \item Dynamic Simulations: For dynamic scenarios, we use excitation methods such as:
    \begin{itemize}
        \item Step inputs: Applying sudden changes in actuation to capture transient behaviors. We show that this method is sufficient for single-actuator cases.
        \item Random actuation (babbling): Implementing variable-hold interval random actuation within a specified range to cover a wide range of dynamic states. This method is appropriate for cases where the system has multiple actuators/inputs. 
    \end{itemize}
\end{enumerate}

These data generation strategies aim to cover a wide range of possible configurations and dynamic behaviors, ensuring that the resulting reduced-order models capture the essential characteristics of each system. We collect snapshots of the system strains (and/or joint twists for rigid systems) for all the static combinations, or at regular intervals throughout the dynamic simulation. The choice of the dynamic simulation duration is determined empirically based on the complexity of the system and the intended application of the reduced-order model. More details and system-specific discussions are presented in the upcoming sections.

\section{Reduction of Soft Manipulators}\label{sec::Analysis}
In this section, we present multiple static and dynamic simulation scenarios to showcase our proposed reduction approach. This analysis focuses on cable-driven soft manipulators with different actuators' paths and number of actuators. The dimensions and material parameters of the manipulators are detailed in Table \ref{tab::ManipulatorParams_6}. These material parameters, which are utilized in this section's simulations, are the default specified in the SoRoSim toolbox, which represent a typical silicone with moderate values for density, Young's modulus, Poisson's ratio, and damping constant. These properties provide a balance of strength, flexibility, and weight appropriate for our modeling purposes. We compare the ROM solution to the High-Order Model (HOM) solution (model used for data generation) using the tip position error metric. As discussed earlier in Section \ref{sec::DynStat}, the static equilibrium is solved using root finding methods, with the straight, undeformed configuration being the initial guess for all the presented cases. As for the dynamics, the differential equation (\ref{eq::gendynamics}) is integrated using various numerical integrators. Both solvers are implemented in the MATLAB toolbox SoRoSim \cite{AnupSoRoSim}. Throughout this Section and Section \ref{sec::POD_Hybrid}, we discuss the reduction of computational cost through both the speed-up factor metric (HOM solution time divided by ROM solution time) and its inverse, the normalized computation time, where the HOM corresponds to a value of 1 for both metrics.

\begin{table}[]
\centering
\caption{\small Parameters of the cable driven soft manipulators.}
\begin{tabular}{ l  c  c }
\hline
 & Single/Three-Actuators & Six-Actuators \\ \hline
Length & $25 \; cm$ & $60 \; cm$ \\ \hline
Base Radius & $1.25 \; cm$ & $2 \; cm$ \\ \hline
Tip Radius  & $0.5 \; cm$ & $1 \; cm$ \\ \hline
Density & \multicolumn{2}{c}{$1000 \; kg/m^3$} \\ \hline
Young's Modulus & \multicolumn{2}{c}{$1 \; MPa$} \\ \hline
Poisson's Ratio & \multicolumn{2}{c}{$0.5$} \\ \hline
Damping Constant & \multicolumn{2}{c}{$10^4 \; Pa \cdot s$} \\ \hline
\end{tabular}
\label{tab::ManipulatorParams_6}
\end{table}

\subsection{Single-Actuator Manipulator}\label{sec:Single cable planar- no gravity}

In this scenario, we introduce a manipulator actuated by a single planar actuator. By "planar" we imply that in an undeformed state, the actuator and the manipulator's centerline \( \boldsymbol{r}(X) \) coexist in a single plane. The actuator path is linear, defined by $\boldsymbol{d}_1(0) = [0,0,1]^T \; cm$  and $\boldsymbol{d}_1(L) = [0,0,0.3]^T \; cm$ at the base and tip of the manipulator, respectively.

\begin{figure}[]
    \centering
    \includegraphics[width=\linewidth]{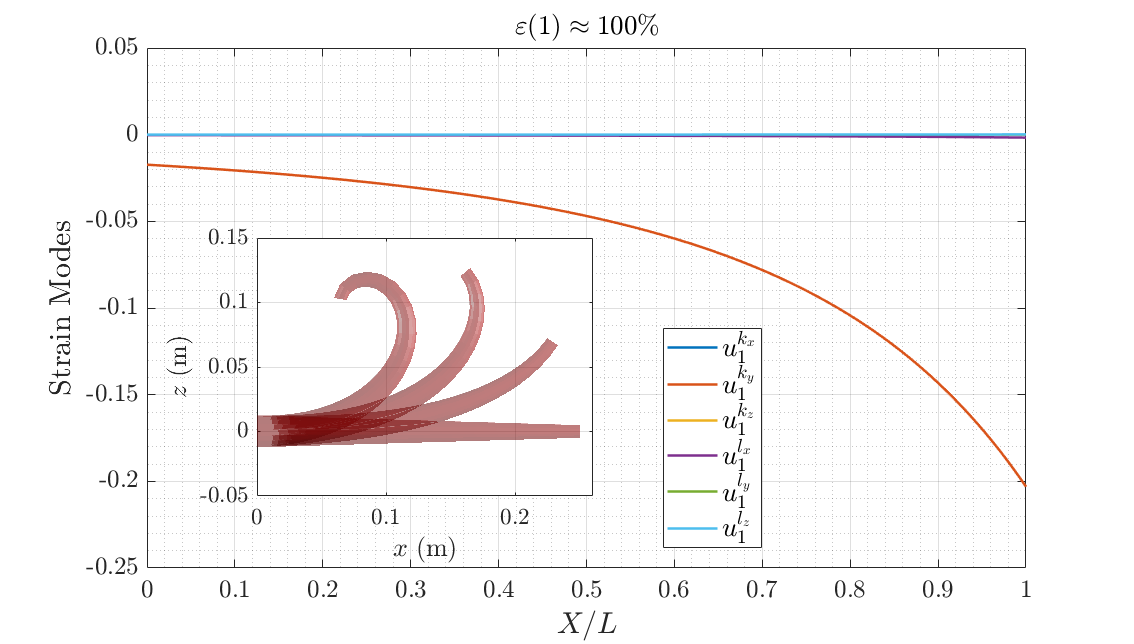}
    \caption{\small The resulting $1^{st}$ mode of the decomposition for the scenario in Section \ref{sec:Single cable planar- no gravity}. As inset, the corresponding family of shapes produced from different scaling of the mode. }
    \label{fig::SingleCabNoGrav_mode}
\end{figure}

\begin{figure*}[ht!]
    \centering
    \includegraphics[width=\linewidth]{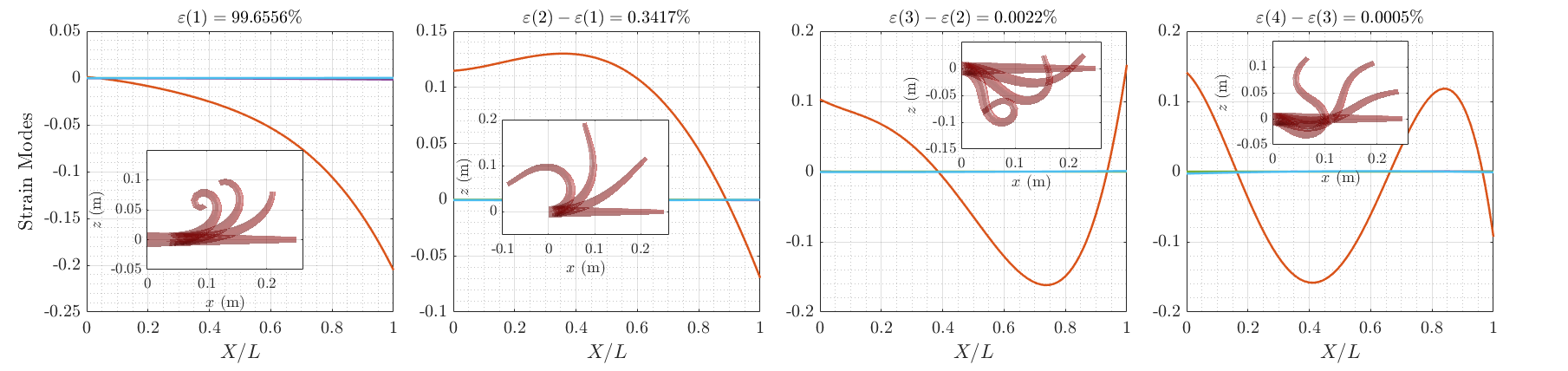}
    \caption{\small The resulting first four modes of the decomposition for the scenario in Section \ref{sec:Single cable planar- no gravity} with dynamics and gravitational load. As inset, the corresponding family of shapes produced from different scalings of each mode. The relative energy is indicated above each mode. The colors follow the same scheme of Figs. \ref{fig::SingleCabNoGrav_mode},\ref{fig::MultiCabNoGrav_Modes} and \ref{fig::6Cables_Mode}. However, only the red (y-bending) is apparent due to the planar actuator path in the $x-z$ local plane, while the other strains are almost zero.}
    \label{fig::SingleCab_modes_DynWGrav}
\end{figure*}

\subsubsection{Statics}


In statics, we define the strain fields (6D) using high-order polynomials. This approach is employed to generate high-fidelity simulation data. In this example, the basis matrix is defined to describe each of the 6 strains with Legendre polynomials up to the $10^{th}$ order, resulting in 66 degrees of freedom. As the computational demand of solving the statics is low, this large number of degrees of freedom are initially defined to converge to the true solution. However, it's worth noting that a smaller number could suffice. For instance, if certain strains are not activated, their corresponding basis terms can be omitted. Similarly, the polynomial order can be lowered if higher-order terms offer no significant advantages. 

The strain snapshots are collected and organized according to the methodology discussed in Section \ref{sec::POD}. Each snapshot is associated with an actuation value ranging from -5 N to 5 N in steps of 0.5 N, resulting in 21 snapshots. Following the execution of the POD technique previously discussed, we examine the resulting decomposition modes. Fig. \ref{fig::SingleCabNoGrav_mode} showcases the first mode obtained. It is apparent that unidirectional bending dominates, with insignificant shear and elongation components. For this principal mode, the contribution is \( \varepsilon(1) \approx 100\% \), indicating that this mode alone can single-handedly describe the system's behavior. This reveals two key points: i) in this specific scenario, a single coordinate is sufficient to describe the system, and ii) all original data snapshots are merely scaling variations of this single mode, enabling its acquisition through a single snapshot instead of multiple ones. 

Owing to the lack of external forces in this specific case, the static equilibrium equation, as represented by Equation (\ref{eqn::genStcEqn}), is reduced to \( \boldsymbol{Kq} = \boldsymbol{B}(\boldsymbol{q}) \boldsymbol{T} \). Generally, the dependence of \( \boldsymbol{B} \) on \( \boldsymbol{q} \) is minimal for planar actuator paths and non-existent for parallel (to the centerline) paths, assuming no external load is applied. This fact explains the presence of a single mode that can fully characterize the system, in agreement with \cite{GeomExct_Renda2022}.


\subsubsection{Dynamics}

Furthermore, we extend our study to the dynamic behavior of the soft manipulator under the influence of gravity. To clearly see what the decomposition modes resulting from the dynamics are, we simulate a step input in this case until steady-state is reached.  Due to the high computational demand of dynamic simulations, we reduced the order of the HOM. We describe the manipulator with a second-order Legendre polynomial for each of the torsional, elongation, and both shearing strains, along with a fourth-order Legendre polynomial for each of the two bending strains. This parametrization results in a total of 22 DOFs.

We find that a single step input with excited transients, in the presence of gravity, results in performance comparable to longer babbling sequences when tested. Thus, we proceed with a single step of 5 N for 1.25 seconds (until the transients die out). This process consumes about 4 seconds of computation. We sample the strain snapshots at 100 Hz, resulting in a total of 126 snapshots. Fig. \ref{fig::SingleCab_modes_DynWGrav} shows the resulting first four main modes along with their relative energy contribution. It can be noted that now the gravitational load along with the transients of the dynamic behaviour introduce small, yet non-trivial modes to the decomposition. In this example, we showed the robot's dynamic modes for the combined inertial (dynamic), actuation, and external loads (gravity). However, the dynamic case without gravity or static case with gravity may require additional modes to fully describe the system under such conditions.
\subsection{Three-Actuator Manipulator}\label{sec::Multi cable}
In this scenario, we explore a case involving multiple actuators. The manipulator under consideration has three actuators, with the same dimensions and properties as Section \ref{sec:Single cable planar- no gravity} scenario, as detailed in Table \ref{tab::ManipulatorParams_6}. The first actuator is exactly the one described previously in Section \ref{sec:Single cable planar- no gravity}. The second is a non-planar, linear actuator, defined by $\boldsymbol{d}_2(0) = [0,1,0]^T \; cm$ and $\boldsymbol{d}_2(L) = [0,0,-0.3]^T \; cm$. The third actuator is aligned parallel to the centerline with $\boldsymbol{d}_3(X) = [0,-0.5,0]^T \; cm$. We employ the same basis definition for data generation as discussed in Section \ref{sec:Single cable planar- no gravity} statics scenario. 

Given that now there are three actuators, the actuation space needs to be sampled comprehensively. This requires dividing the actuation space into equal increments and testing all possible combinations, resulting in exponential complexity. In this specific case, we use 11 steps within the \([-5, 5]\) N range, yielding a total of \(11^3 = 1331\) snapshots. 

\begin{figure}[]
    \centering
    \includegraphics[width=\linewidth]{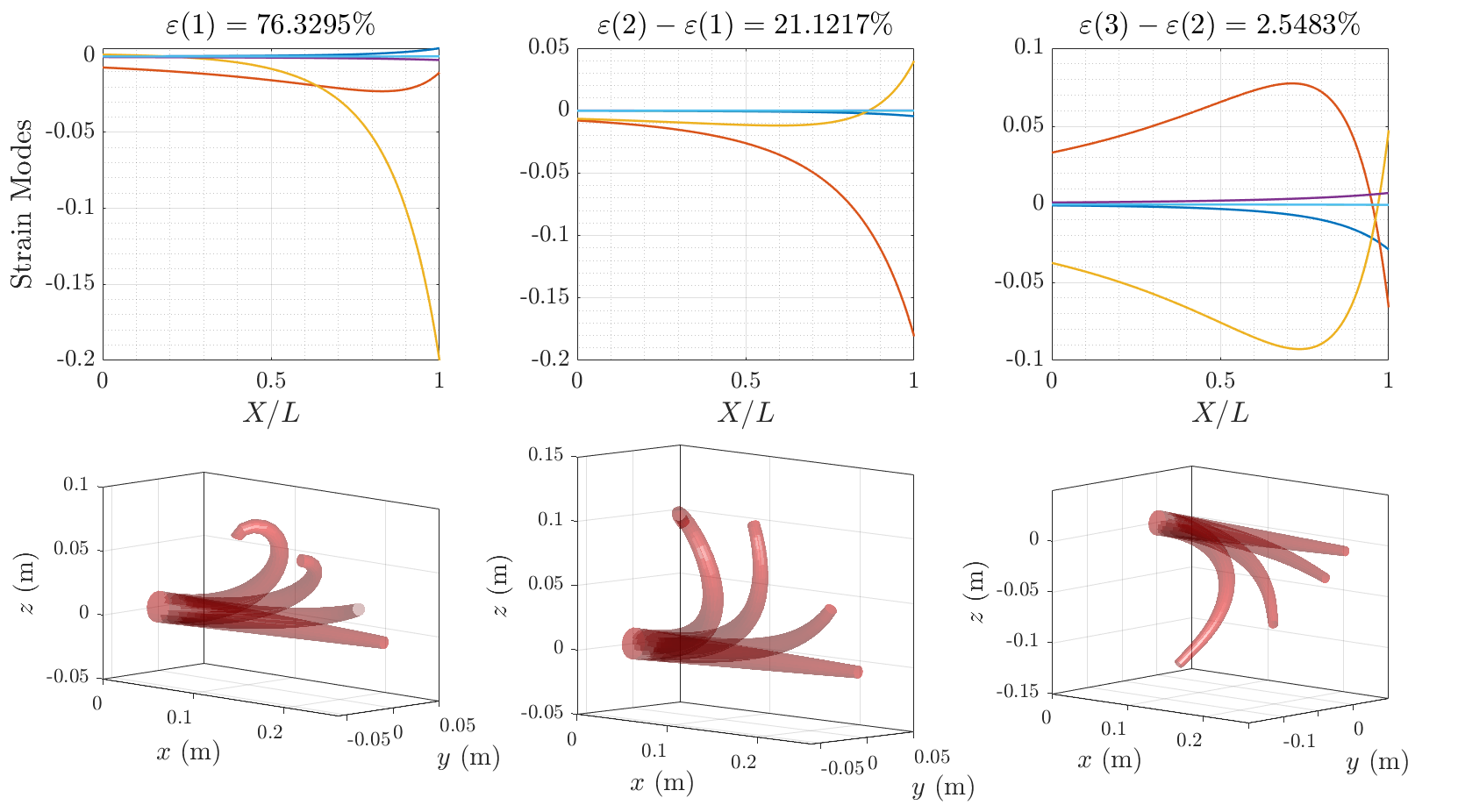}
    \caption{\small The first three modes of Section \ref{sec::Multi cable} scenario decomposition, with their respective family of shapes below. The relative energy is indicated above each mode. The blue, red, yellow, purple, green and light blue curves are the modes of the torsional, y-bending, z-bending, elongation, y-shearing and z-shearing strains respectively.}
    \label{fig::MultiCabNoGrav_Modes}
\end{figure}

Upon performing the decomposition, a set of modes is generated that collectively represent the system. The main three primary modes, whose sum of relative energies are almost 100\%, are depicted in Fig. \ref{fig::MultiCabNoGrav_Modes}, highlighting that all rotational strains now contribute to the deformation of the manipulator. 

\begin{figure*}[ht!]
    \centering
    \includegraphics[width=\linewidth]{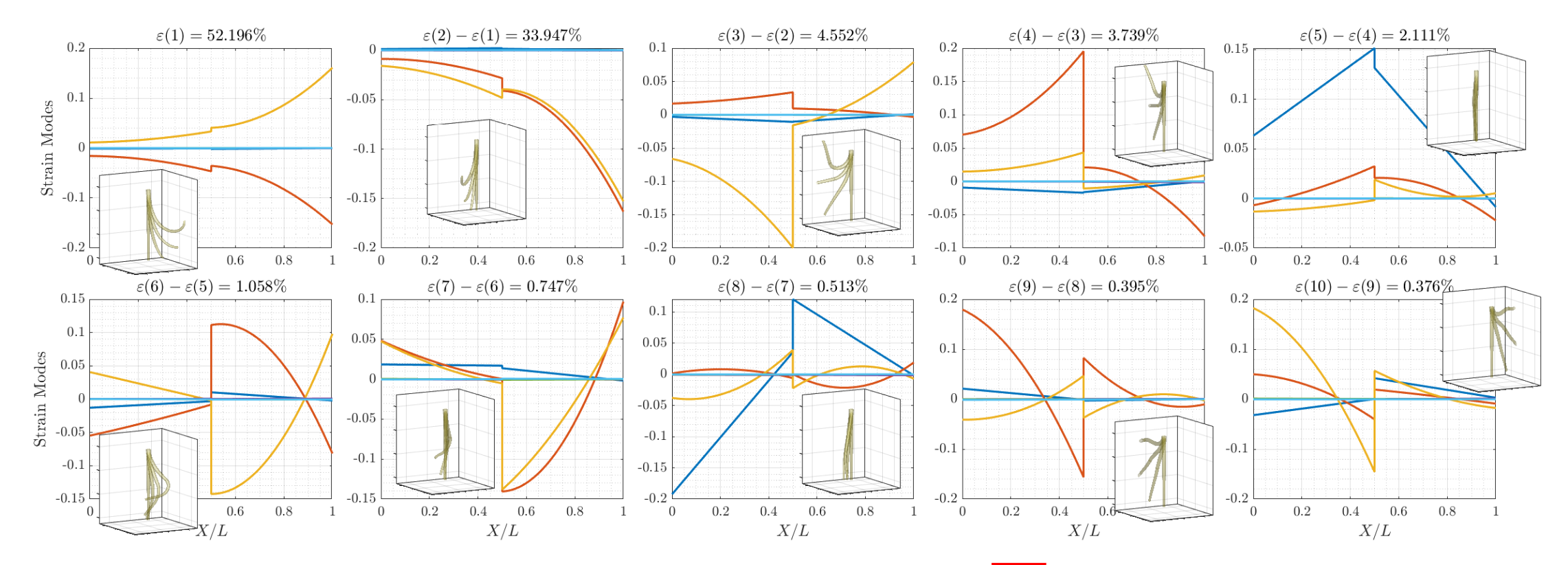}
    \caption{\small The resulting first ten modes of the decomposition for the scenario in Section \ref{sec::6Act}. As inset, the corresponding family of shapes produced from different scaling of each mode. The relative energy is indicated above each mode. It is observed that the planes of bending in mode 1 and 2 are perpendicular, and so is the case with mode 3 and 4. The blue, red, yellow, purple, green and light blue curves are the modes of the torsional, y-bending, z-bending, elongation, y-shearing and z-shearing strains respectively.}
    \label{fig::6Cables_Mode}
\end{figure*}
\subsection{Six-Actuator Multisection Manipulator}\label{sec::6Act}

In this section, we present a six-actuator manipulator where the actuation domain of three span the whole body while the other three span only half of it. The actuators are placed 60 degrees apart and alternating in their actuation domain, their path is straight, and at the surface of the manipulator. The manipulator is positioned vertically and subjected to gravitational load, in contrast to the cantilever position presented earlier. The dimensions and material properties of the manipulator can be seen in Table \ref{tab::ManipulatorParams_6}.

The cables affecting only half of the manipulator introduce strain discontinuities at the terminal point. These discontinuities can affect solutions approximated by continuous functions. To address this, while defining the HOM, we split the abscissa domain $X$ into two sections, assigning separate strain bases to each. This approach effectively captures the discontinuity. For each manipulator section, we use a first-order Legendre polynomial (linear) for torsional, elongation, and shear strains, and a second-order Legendre polynomial (quadratic) for bending strains. This results in 28 DOFs for the entire manipulator. Note that a discontinuous basis covering the whole manipulator, with non-zero values only at either sections, can be used equivalently. 

Considering the presence of multiple actuators, exploring the actuation space requires a different approach. In statics, this involves dividing the actuation space into equal segments and evaluating all combinations, leading to exponential complexity. However, we shift our attention now, presenting an appropriate method to cover the actuation space for dynamics. We find that babbling is most effective in covering the actuation space. Babbling is generally the process of repeatedly executing a randomized actuation signal for some period of time. It can be implemented in various ways, however, we do so by holding a random actuation value within a specific range for a specific time per actuator. To further improve on this, we do not fix the holding time; instead, we vary it randomly within a chosen interval. This way, the number of possible combinations increases, maximizing the actuation space coverage. For our case, we actuate the manipulator within the range $[-20 \; 20] \; N$, and the holding time is within $[0.5 \; 1] \; s$, changing independently for each actuator. 
We simulate 1 minute of variable-hold babbling, sampled at 100 Hz. Upon gathering and arranging the 6k strain snapshots, we perform the POD. The resulting decomposition modes, their respective family of shapes and relative energy can be seen in Fig. \ref{fig::6Cables_Mode}. The benefit of dividing the domain into multiple sections can already be seen from the resulting modes. That is because the jump in strain introduced by the shorter actuators would have been tough to capture using a single section. In addition, the resulting optimal mode is discontinuous, and treats the whole manipulator as one section, further reducing the required DOFs. Accommodating discontinuities in strain fields is one of the advantages over position-based models, as discussed in \cite{GeomExct_Renda2022}. It is worth noting that the first two modes generally represent the deformation caused by the longer actuators. The planes of deformation for these two modes are orthogonal, thus linear combinations of both modes can rotate the plane of deformation. As for the following two modes, they correspond to the shorter actuators, and the deformation planes are orthogonal as well. These present a good example of embodied orthogonality seen from the POD. The higher modes correspond to the remaining motions induced by the transients and gravitational load.
\subsection{ROM Validation and Comparison}

While in previous sections we performed our reduction approach, analyzed the results, and visualized the decomposition modes, we never used these modes to parameterize the strain fields and solve the statics or dynamics of the system. In this section, the performance, computational load and the interpolation-extrapolation abilities of the new ROM are presented. Our analysis aims to utilize minimal data, subsequently comparing the ROM, varying in the number of degrees of freedom, with the HOM.

\begin{figure}[]
    \centering
    \includegraphics[width=\linewidth]{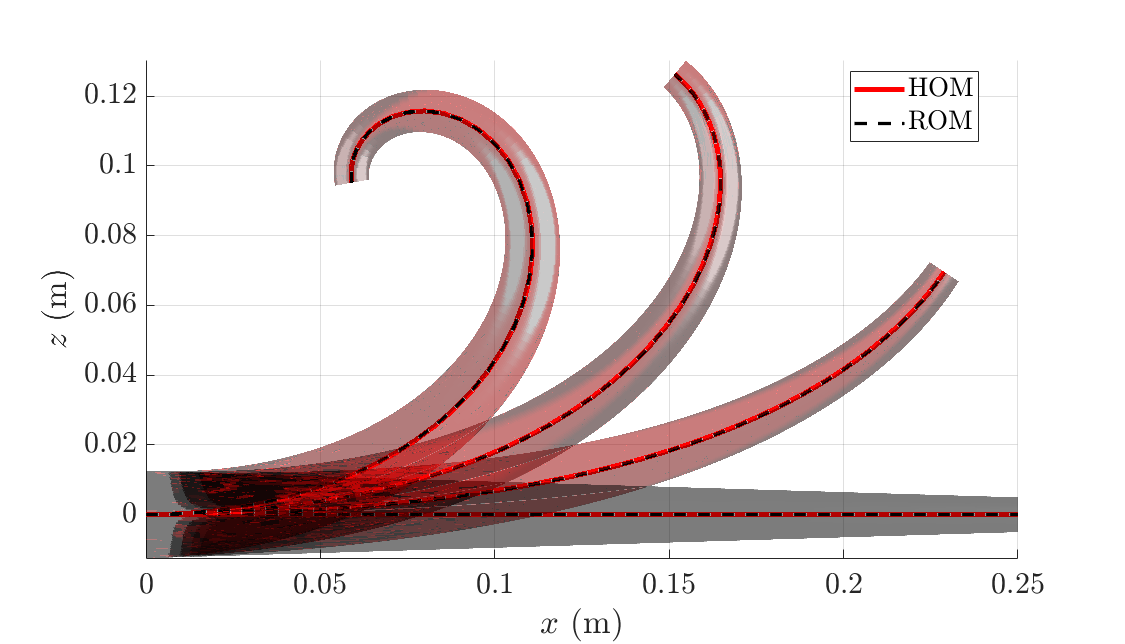}
    \caption{\small Comparison between the HOM and ROM solutions for the scenario in Section \ref{sec:Single cable planar- no gravity}.}
    \label{fig::SingleCabNoGrav_config}
\end{figure}
\begin{figure}[]
    \centering
    \includegraphics[width=\linewidth]{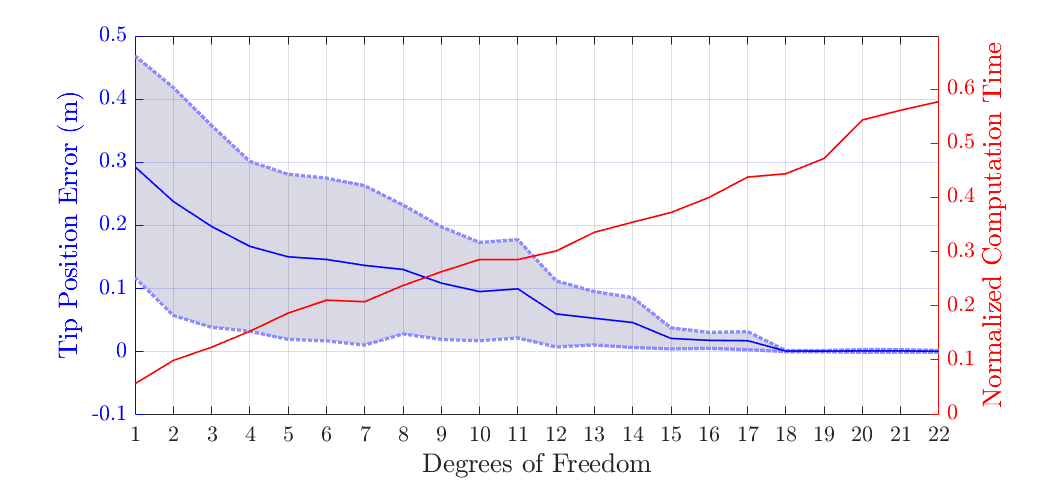}
    \caption{\small Tip position error (between the ROM and HOM) and normalized computation time against the number of DOFs for the ROM for the scenario in Section \ref{sec::6Act}. The average (solid) and the standard deviation (envelope) of the tip position error along the 10 seconds of testing are shown in the shaded error envelope. The error converges to almost zero starting from 18 DOFs, marking a clear choice for the number of DOFs.}
    \label{fig::DOF_VS_ErrTime}
\end{figure}
\begin{figure}[]
    \centering
    \includegraphics[width=\linewidth]{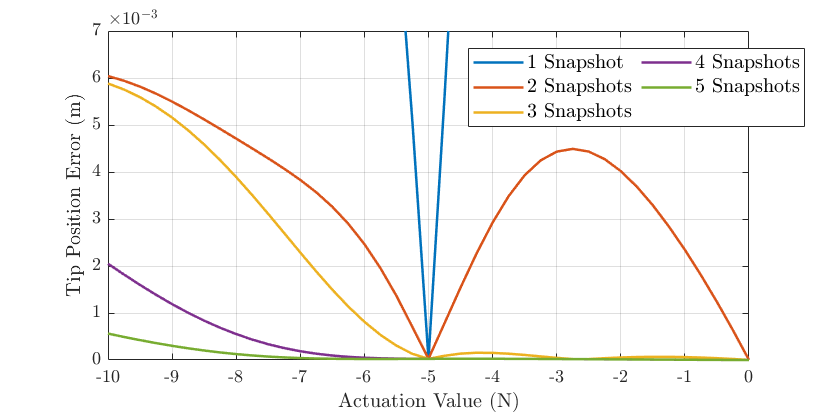}
    \caption{\small Interpolation-extrapolation abilities for multiple cases of different snapshots. The error axis is limited to what is presented due to the difference in the order of magnitude for the single snapshot case (blue), where the error reaches 0.092 and 0.045 m at 0 and -10 N respectively.}
    \label{fig::IntExtFig}
\end{figure}

\subsubsection{Single-Actuator Manipulator}
Utilizing the single obtained coupled mode in the static, zero-gravity analysis in Section \ref{sec:Single cable planar- no gravity}, we solve the equilibrium equation for multiple actuation values that fall inside and outside the original data range. The resulting configuration is then compared with the HOM, as depicted in Fig. \ref{fig::SingleCabNoGrav_config}. We can see an almost perfect match between both configurations, with tip position errors of $(0,11.96,34.33,44.21) \; \mu m$  at the actuation levels of $(0,-2,-5,-10)$ N respectively. It is worth noting that using a single snapshot for the POD (i.e. the snapshot is the mode), the tip errors are almost unchanged, confirming our earlier remark. In addition, we assess the computational cost reduction for the dynamic case with gravity, whose modes are shown in Fig. \ref{fig::SingleCab_modes_DynWGrav}. It is observed that simulating 10 seconds of continuously changing actuation required only 2 seconds of computation while using all 4 main modes, allowing for utilization in real-time applications.

\subsubsection{Three-Actuator Manipulator}

To validate the ROM, the least number of modes is employed to solve the statics of Section \ref{sec::Multi cable} system. It is found that a minimum of three modes is necessary to achieve a configuration that closely resembles the HOM, as the solution deviates significantly when using one or two modes. We conclude, as anticipated, there exists a lower limit on the number of modes needed for a reliable reduced-order representation. Generally, the number of modes should be at least equal to the number of actuators, i.e. \(n \geq n_a\). Using these three modes, the system's static equilibrium is then solved for multiple actuation levels, within and beyond that of the original data, similar to what was done in earlier. The resulting configurations match almost perfectly with the HOM solutions, exhibiting errors in the order of micrometers. Multiple comparison cases can be seen in the supplementary video for this scenario, as well as the single actuator case, where we report the RMSE metric using multiple points along the manipulators body.

\subsubsection{Six-Actuator Multisection Manipulator}

We test our ROM in dynamics for the scenario presented in Section \ref{sec::6Act}. A random discrete signal of 1 Hz is applied to each actuator for 10 seconds. Then, the tip position of the ROM solution is compared to that of the original HOM. The range of actuation is 50\% larger than that used for the HOM. The error between both tip positions can be seen in Fig. \ref{fig::DOF_VS_ErrTime} when using various numbers of DOFs. It can be seen that, as expected, the error monotonically decreases as the degrees of freedom increase. At 18 DOFs, the error drops to almost zero, showing the applicability of the ROM to actuation ranges beyond the original. 

As previously discussed, some strains are usually of different orders of magnitude than others. This issue of scale and mode representation reflects the results of our analysis of the relative energy content across different modes, as illustrated in Fig. \ref{fig::6Cables_Mode}. The cumulative relative energy content, such as \(\varepsilon(10) = 99.634\%\), might suggest comprehensive model behavior coverage. However, as demonstrated by the substantial positional error observed at this level of truncation in Fig. \ref{fig::DOF_VS_ErrTime}, relying solely on cumulative energy content can be misleading. The existence of specific higher modes, although contributing slightly to the total energy, is critical for capturing small yet crucial deformations essential for accurately describing the dynamics of the system.

More system-specific metrics, such as the position error metric, are necessary for a more accurate and functional truncation method in the modeling process. This ensures that all relevant dynamical features are preserved, especially those that are critical yet not immediately apparent from standard energy content analyses.

\subsubsection{Interpolation-Extrapolation Performance}

To demonstrate the effect of the number of snapshots and the interpolation-extrapolation abilities (i.e., the generalization capabilities given few data points within and beyond the actuation range), we study the statics of the single-actuator manipulator described in Section \ref{sec:Single cable planar- no gravity} with the inclusion of gravity. We chose this case due to the apparent non-linearity present due to the gravitational load, that introduce additional non-trivial modes to the decomposition. In the study, we solve the static equilibrium of the system under a range of actuation, specifically [-10 0] N, for five different cases. Each case corresponds to a different number of snapshots used to acquire the optimal strain bases used to solve the system. The snapshots are taken equidistantly within the [-5 0] N range. For instance, the three snapshots case utilizes the strain acquired from 0, -2.5 and -5 N of actuation. The special case of a single snapshot uses just one actuation value at -5 N. The error in tip position between the resulting ROM and the HOM is shown in Fig. \ref{fig::IntExtFig} for the different cases. It is intuitive that as the number of snapshots increases, the interpolation capabilities are better, and this is clearly seen from the figure. However, what is remarkable here is the extrapolation abilities that are drastically improved over a wide range outside the snapshots' domain, when more snapshots are used. It is worth noting here that we use as many modes as snapshots for each presented case. While this would make the comparison a bit unfair, the insights gained from the study are valuable, and the same would apply for the cases of multiple actuators and snapshots.
\section{Reduction of Hybrid Soft-Rigid Robots}
This section demonstrates the use of our POD-based ROM technique in contexts beyond slender soft manipulators. We aim to illustrate the versatility and effectiveness of this approach in a variety of robotic systems, including hyper-redundant rigid robots and hybrid closed-chain prototypes.
\subsection{Cable-Driven Hyper-Redundant Manipulator}\label{sec::Spiral}

\begin{figure}[]
    \centering
    \includegraphics[width=\linewidth]{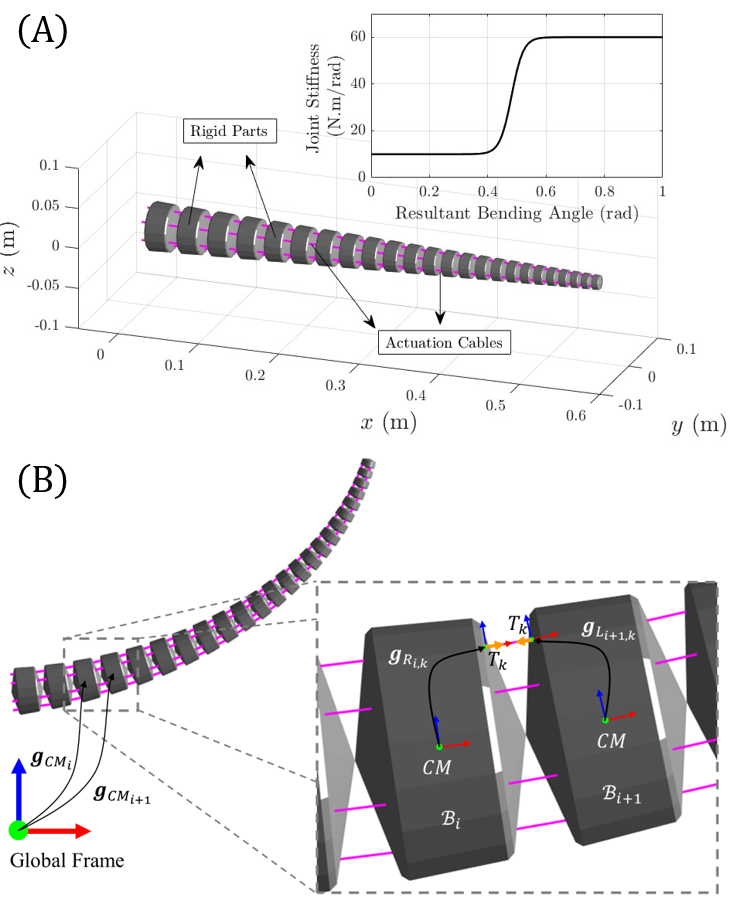}
    \caption{\small Schematic of the cable-driven hyper-redundant manipulator. (A) 24 rigid bodies are connected to each other through spherical joints with a nonlinear stiffness model, as shown in the inset. (B) Actuation model}
    \label{fig::FBD_Spiral}
\end{figure}

The first example we apply our approach to is a robot that resembles a soft, slender manipulator but consists of 24 rigid parts connected by spherical joints, resulting in a total of 72 DOFs. The manipulator is driven by six cables, separated by 60 degrees, and alternating in their actuation domain, similar to that presented in Section \ref{sec::6Act}. The spherical joints have non-linear stiffness that rises significantly at a specific angle (governed by the geometry), modeling the contact between the rigid parts. A system diagram is presented in Fig. \ref{fig::FBD_Spiral}.

\begin{figure*}[ht!]
    \centering
    \includegraphics[width=\linewidth]{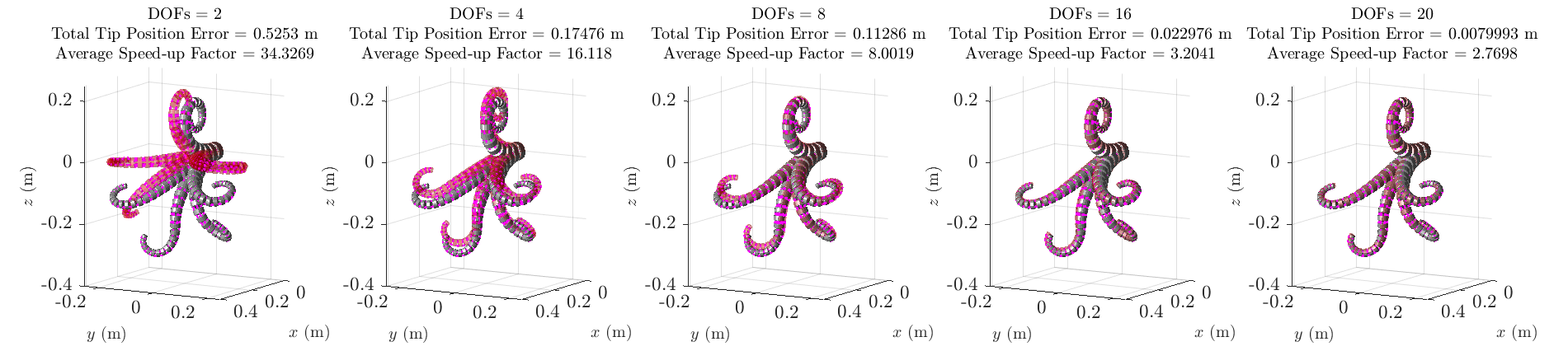}
    \caption{\small Solutions comparison between the HOM and the ROM of the cable-driven multi-joint manipulator. Each panel has the solution of five different actuation combinations of the six cables for the 72 DOFs HOM (opaque gray body) as well as the ROM (transparent red body). Table \ref{tab::SpiralActLevels} details the actuation values used. The total tip position error is the sum of each corresponding tip position error for the five actuation levels. The solution time speedup factor is calculated for each case, and the average is reported. }
    \label{fig::SpiralSnapshots}
\end{figure*}
The actuation force of the cables is computed according to Fig. \ref{fig::FBD_Spiral}B. Consider $k^{th}$ cable and adjacent rigid bodies $i$ and $i+1$. A fixed transformation matrix from the center of mass (CM) of the body $i$ to the cable outlet on the right side is given by ${\bm{g}_{R}}_{i,k}$. Similarly, ${\bm{g}_{L}}_{i+1,k}$ is the fixed transformation matrix that connects the CM of the body $i+1$ to the cable outlet on its left side. Hence, the transformation matrix from the global frame to the location of $k^{th}$ cable on each body is given by ${\bm{g}_{CM}}_i{\bm{g}_{R}}_{i,k}$ and ${\bm{g}_{CM}}_{i+1}{\bm{g}_{L}}_{i+1,k}$. Using this, we can compute the direction of the actuation force on body $i$ (${\bm{T}_{R}}_{i,k}$) and $i+1$ (${\bm{T}_{L}}_{i+1,k}$). Every rigid body, except the terminal one, is subject to cable tension on the left and right sides. The combined actuation wrench on body $i$ is given by,

\begin{equation}
    {\boldsymbol{\mathcal{F}}}_{a_i} = \Sigma_{k=1}^{n_a} \left( \text{Ad}^*_{{\bm{g}_{L}}_{i,k}}[\bm{0} \; {\bm{T}_{L}^T}_{i,k}]^T+\text{Ad}^*_{{\bm{g}_{R}}_{i,k}}[\bm{0} \; {\bm{T}_{R}^T}_{i,k}]^T \right).
\end{equation}

In statics, we apply only 0 and -25 N for each cable, resulting in 64 snapshots. Those are basically the corners of the 6D hypercube in the actuation space. The snapshots of the solution -in this case angles- are then arranged and decomposed as discussed in Section \ref{sec::POD}. The resulting modes of decomposition are then used to solve the ROM. 

Upon finding the decomposition matrices, the first $n$ left singular vectors are used as the reduced system basis matrix ${\boldsymbol{\Phi}_{\bar{\bar{\xi}}_{\mathcal{O}}}} \in \mathbb{R}^{(24\times6) \times n}$. This reduced basis matrix maps the reduced generalized coordinates $\boldsymbol{q}$ to the HOM joint twist $\boldsymbol{\xi}_i$ as follows:

\begin{equation}
    \boldsymbol{\xi}_i = \bm{\mathcal{E}}_i{\boldsymbol{\Phi}_{\bar{\bar{\xi}}_{\mathcal{O}}}}\boldsymbol{q}
\end{equation}
where, $\bm{\xi}_i = [\theta_x,\;\theta_y,\;\theta_z,\;0,\;0,\;0]^T_i$. The resulting modes now couple all rotations of the spherical joint for all joints, rather than each direction of each joint being governed by a separate coordinate. This clearly shows how our approach extends the concept of robot synergies to spatial coupling and multi-directional strains/twists.

The ROM is then solved and compared to the HOM. For different numbers of DOFs, both solutions are compared at 5 different actuation levels, outlined in Table \ref{tab::SpiralActLevels}. The resulting configurations, along with errors and computation speed-up factors, are shown in Fig. \ref{fig::SpiralSnapshots}. It can be observed that at 16 DOFs, the total tip position error is minimal, and the configurations of the HOM and ROM solutions are indistinguishable. This is a significant reduction from the original 72 DOFs, while exhibiting negligible errors and speeding the computational time needed by a factor of 3.2.

\subsection{Soft-Legged Platform}\label{sec::SLP}

\begin{figure}[]
    \centering
    \includegraphics[width=0.9\linewidth]{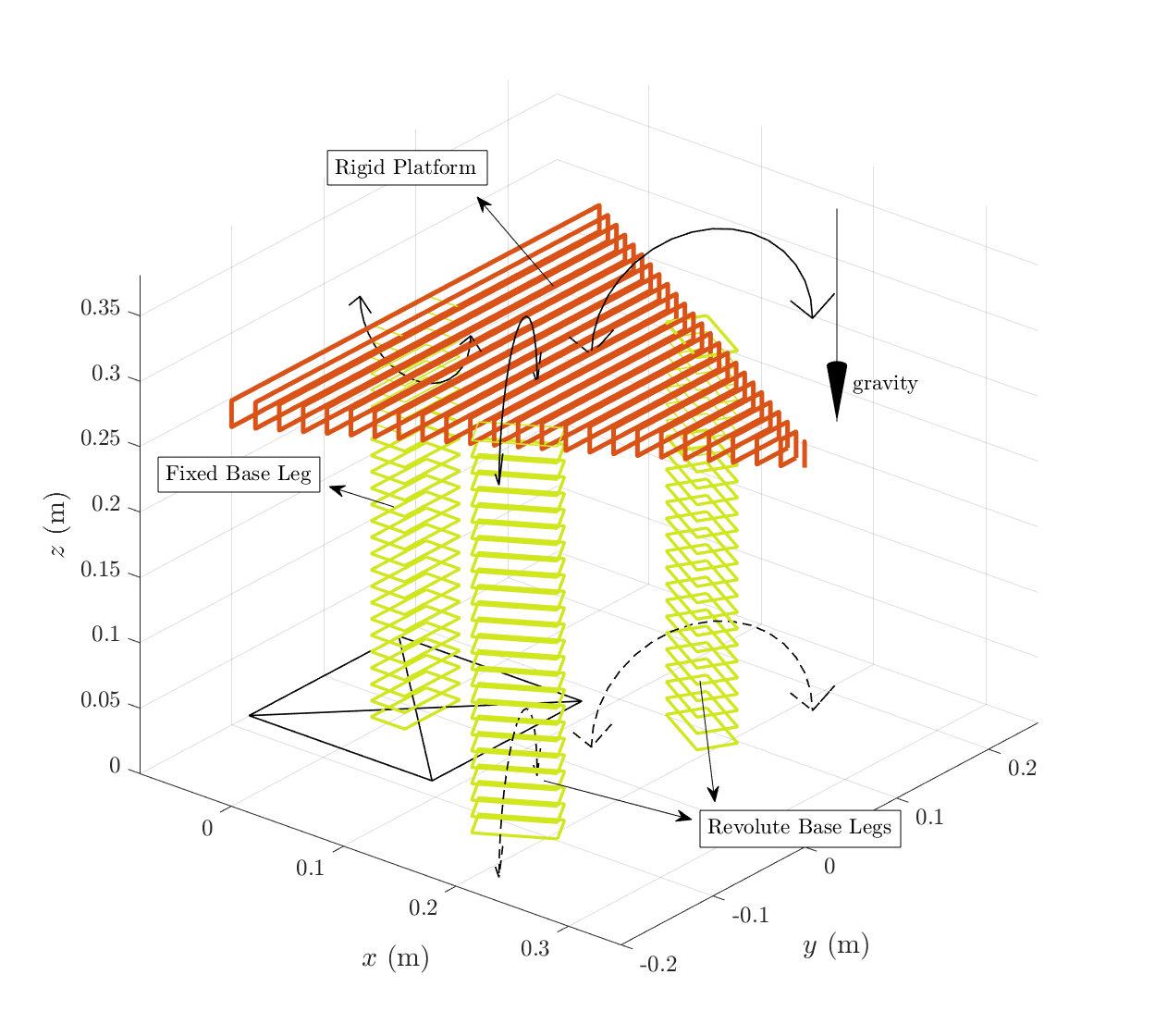}
    \caption{\small Schematic of the closed chain soft-legged platform. The crossed square represents a fixed joint, while the semicircular arrows represent revolute joints. The dashed semicircular arrows are closed-loop revolute joints.}
    \label{fig::FBD_SLP}
\end{figure}

So far, we have applied our proposed ROM method to entirely soft, strain-parametrized prototypes (Section \ref{sec::Analysis}), as well as an entirely discrete-jointed prototype (Section \ref{sec::Spiral}). In this section, we test the method on a hybrid system that consists of both soft and rigid components with discrete joints. The system consists of a triangular ($40\; cm$ isosceles), rigid platform with three soft legs ($30\; cm$ with rectangular $6\times 3 \; cm^2$ cross-section) connected to the center of each triangle edge through a revolute joint. Two of the legs are connected to the ground with revolute joints, while the third is connected with a fixed joint for a restoring force, making the resulting system a hybrid, closed-chain prototype. A diagram of the system can be shown in Fig. \ref{fig::FBD_SLP}.

Each soft leg is parameterized with a third-order Legendre polynomial for each of the 6 strains, i.e. 18 DOFs per leg. With three degrees of freedom for the revolute joints, the total becomes 57 DOFs. The remaining two revolute joints are closed-loop joints and are not counted as degrees of freedom. The prototype is excited by a random 6D wrench applied at the CM of the rigid platform. The wrench changes every 0.25 to 0.5 seconds, and the system is simulated for 50 seconds. Snapshots of the system response are then gathered and arranged as discussed in Section \ref{sec::POD_Hybrid}. Upon performing the POD, we acquire bases that encompass both the soft (strain coefficients) and rigid (joint angles) coordinates. The new bases are used to solve the system.
\begin{table}[]
\centering
\caption{\small The five actuation levels for the cable-driven hyper-redundant manipulator comparison cases shown in Fig. \ref{fig::SpiralSnapshots}.}
\begin{tabular}{cccccc}
\multirow{2}{*}{} & \multicolumn{5}{c}{Actuation Level (N)} \\ 
                  & 1       & 2     & 3       & 4    & 5    \\ \hline
Cable 1           & -12.5    & -50    & -12.5    & -25   & -25   \\ \hline
Cable 2           & -12.5    & 0     & -12.5    & 0    & 0    \\ \hline
Cable 3           & 0       & 0     & 0       & -10   & -30   \\ \hline
Cable 4           & 0       & 0     & 0       & 0    & -10   \\ \hline
Cable 5           & 0       & 0     & -25      & -40   & -10   \\ \hline
Cable 6           & 0       & 0     & -10      & -10   & 0    \\ \hline
\end{tabular}
\label{tab::SpiralActLevels}
\end{table}
\begin{figure*}[]
    \centering
    \includegraphics[width=\linewidth]{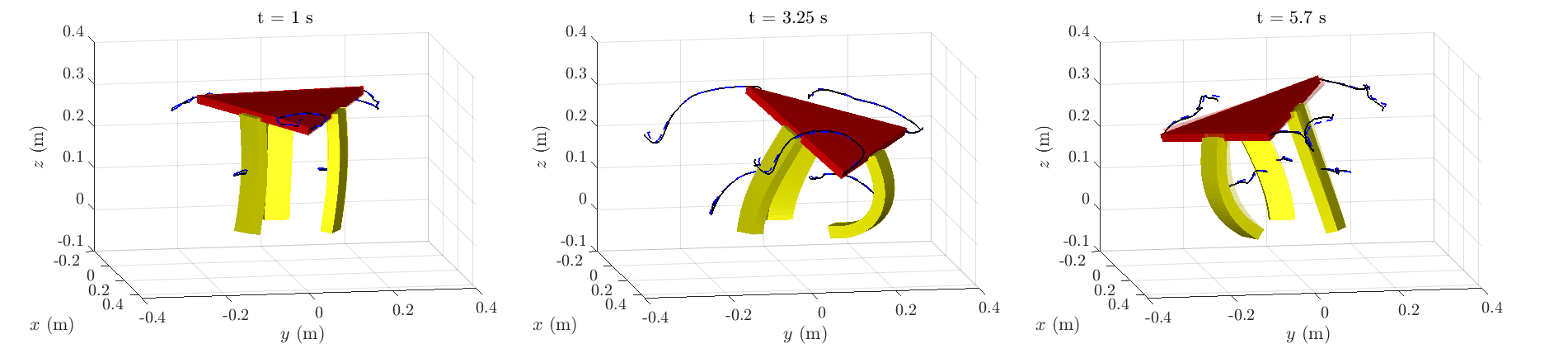}
    \caption{\small Solution snapshots comparison for the 10 s testing between the 57 DOFs HOM and the 25 DOFs ROM of the soft-legged platform. Both HOM and ROM configurations are overlaid in each panel, with the ROM being the transparent one (which can be clearly seen in the third panel only). The trail of the chosen error points over the preceding second are plotted for both HOM (solid black) and ROM (dashed blue) solutions.}
    \label{fig::SLPSnapshots}
\end{figure*}
\begin{figure}[]
    \centering
    \includegraphics[width=\linewidth]{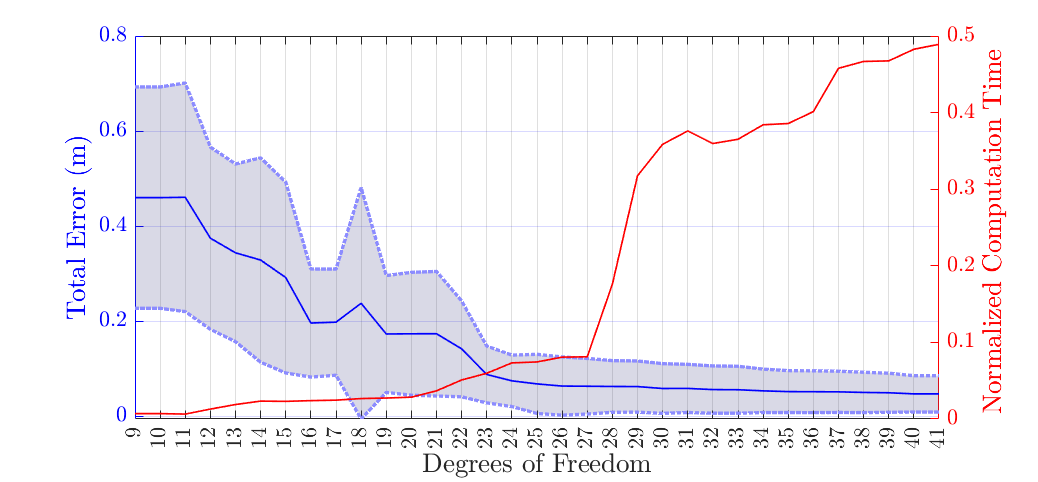}
    \caption{\small Total error (between the ROM and HOM) and normalized computation time against the number of DOFs for the ROM of the soft-legged platform (Section \ref{sec::SLP}). The total error is the sum of the errors of six points, the midpoint of each soft leg, and the tips of the triangular platform. The average (solid) and the standard deviation (envelope) of the total error along the 10 seconds of testing are shown in the shaded error envelope. The modes up to 10 DOFs did not allow for any reaction to the input, keeping the error constant, hence not showing most of them.}
    \label{fig::DOF_VS_ErrTime_SLP}
\end{figure}
We simulate 10 seconds of a new random wrench excitation within the same range using different DOFs for the ROM. To assess the performance of the ROM, we choose the error to be the summation of six points, the midpoint of each soft leg, and
the tips of the triangular platform. The average and standard deviation of the total error over the 10 seconds are reported in Fig. \ref{fig::DOF_VS_ErrTime_SLP}. In the same figure, the normalized computation time (inverse of the speed-up factor) is shown as well. A clear decision about the number of DOFs for the ROM can be made from this analysis, as 23-27 DOFs hold the best error-computation time balance. In addition to using less than half the HOM DOFs, using this range of DOFs results in a speed-up factor of more than 13. The configurations of both HOM and ROM solutions using 25 DOFs at different instants are shown in Fig. \ref{fig::SLPSnapshots}. In addition, the trail of the six chosen error points over the preceding second is compared, showing the close matching of both solutions.

\begin{table}
\centering
\caption{\small Speed-up factors for different scenarios that correspond to ROMs with 5\% tip position error (normalized by one body length). For the soft-legged platform (Section \ref{sec::SLP}), the average of the six chosen points is used, and the length of one leg is used as a body length.}

\begin{tabular}{ccc}

Scenario Section & Simulation type & Speed-up factor \\ \hline
\ref{sec:Single cable planar- no gravity} & Dynamic & $5.2$              \\ \hline
\ref{sec::Multi cable} & Static & $6.8$              \\ \hline
\ref{sec::6Act} & Dynamic & $2.7$              \\ \hline
\ref{sec::Spiral} & Static & $12.6$             \\ \hline
\ref{sec::SLP} & Dynamic & $16.8$              \\ \hline
\end{tabular}

\label{table::SpeedupTable}
\end{table}


\section{Experimental Results}
We applied the ROM formulation to two experimental scenarios. The first experiment focuses on using the ROM for shape estimation in a cable-driven soft robot with minimal position tracking data. In the second experiment, we validate and compare the ROM for a parallel hybrid soft-rigid robot prototype.
\subsection{Application to Shape Estimation}\label{sec::ShapeEstimation}

To evaluate the potential benefits of employing the proposed method for high-dimensional soft robots, we utilize our reduction approach in the context of shape estimation. We achieve this through: (1) modeling and simulating a complex six-cable soft manipulator; (2) obtaining the strain modes capable of efficiently describing the system by decomposing the simulation data; and then (3) utilizing the extracted modes and decomposition insights in a shape estimation task using a minimal number of sensory inputs. This process allows us to verify and demonstrate the applicability of our approach.

In recent years, there has been increasing interest in estimating the state of continuum robots using sensor data. For instance, in  \cite{anderson2017continuum}, \cite{lilge2022continuum} and \cite{ferguson_unified_2024},  the pose (position and orientation) of certain markers is utilized with an electromagnetic tracking system for shape estimation. ROMs hold promise in enhancing shape estimation methods by significantly reducing the variables required to describe the 3D geometry of soft robots. This reduction not only minimizes the sensor requirements but also lowers the computational costs of the algorithms involved. Here, we propose an estimation approach that predicts the shape of soft robots based solely on the measured positions of a limited number of markers. 
Our method differs from other approaches in that it only tracks the positions of markers placed along the soft robot, rather than measuring both position and orientation. This simplifies sensor placement requirements to some extent.

\begin{figure}[]
    \centering
    \includegraphics[width=0.8\linewidth]{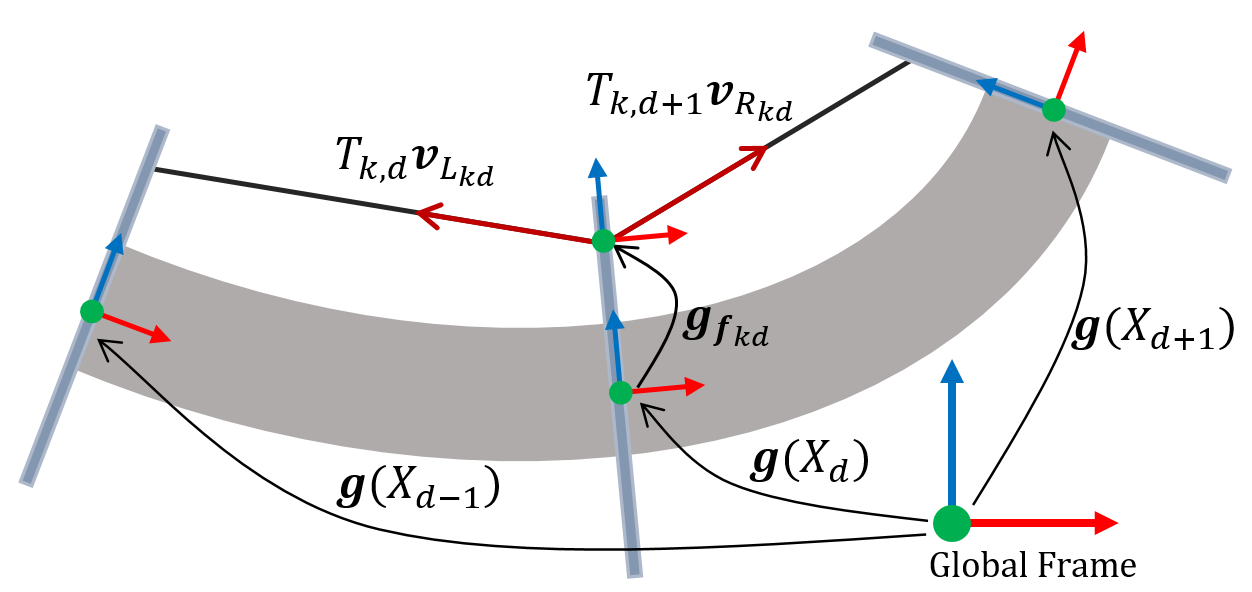}
    \caption{\small Externally disk-guided actuation model. The transformation $\boldsymbol{g}_{\boldsymbol{f}_{kd}}$ is from the local robot frame at disk $d$ location $X_{d}$, to the contact point between the $k^{th}$ actuator and the disk through translation only ($\boldsymbol{R}_{\boldsymbol{f}_{kd}} = \boldsymbol{I}$ and $\boldsymbol{r}_{kd} = \boldsymbol{d}_k(X_{d}$)).}
    \label{fig::Friction}
\end{figure}

We use a silicone-based soft robot prototype with dimensions matching those of the single-actuator manipulator described in Table \ref{tab::ManipulatorParams_6}. The actuation cables are routed externally through disk-guides, which are 3D printed with polylactic acid (PLA), and integrated with the silicone body. The radius of these disks is 3 mm larger than the manipulator's radius at their respective locations. The manipulator is equipped with 6 linearly independent actuators, with 3 of them extending from the base to the tip of the manipulator, while the remaining 3 terminate at the midpoint of the manipulator. In this case, accurately modeling tension loss due to friction at the contact points is crucial. To quantify this frictional effect, we employ the Capstan equation \cite{rone_continuum_2014}:
\begin{equation}
    \frac{T_{k,d+1}}{T_{k,d}} = e^{-\mu \phi_{kd}},
\end{equation}
where:
\begin{equation}
    \phi_{kd} = \cos^{-1}(-\boldsymbol{v}_{L_{kd}}^T\boldsymbol{v}_{R_{kd}}),
\end{equation}
$\mu$ is the coefficient of friction, and $k$ is the actuator index and $\boldsymbol{v}_{L_{kd}},\boldsymbol{v}_{R_{kd}}$ are unit vectors in the $k^{th}$ actuator directions on both sides of the $d^{th}$ guiding disk (see Fig. \ref{fig::Friction}). The resultant actuation force $\boldsymbol{f}_{kd}$ is:
\begin{equation}
    \boldsymbol{f}_{kd} = \boldsymbol{v}_{L_{kd}}T_{k,d}+\boldsymbol{v}_{R_{kd}}T_{k,d+1},
\end{equation}

\noindent which can then be considered a concentrated force at the cable hole. The corresponding actuation wrench acting on disk center $d$ is given by:


\begin{equation}
\small \boldsymbol{\mathcal{F}_{a_d}}(X) = \sum_{k=1}^{n_a}  \text{Ad}^*_{\boldsymbol{g}_{\boldsymbol{f}_{kd}}} \left[\begin{array}{c} \bf{0} \\
\boldsymbol{v}_{L_{kd}} + \boldsymbol{v}_{R_{kd}}e^{-\mu \phi_{kd}}\end{array}\right] e^{-\mu\sum_{j=1}^{d-1}\phi_{kj}}{T}_{k}
\end{equation}

\noindent where $n_d$ is the number of guiding disks. The wrench can then be projected onto the generalized coordinates space through the Jacobian at the point of application: soft body centerline. The material parameters for the prototype were identified through initial experiments involving varying actuation values and external loads. Through these experiments, we accurately determined the density, Young's modulus, and coefficient of friction. The identification process, which used measurements from a motion capture system and an optimization algorithm, matched the model with experimental data, yielding a Young's modulus of \(0.75 \; \text{MPa}\), a density of \(1349.5 \; \text{kg/m}^3\), and a friction coefficient of \(0.35\).

\begin{figure*}[]
    \centering
    \includegraphics[width=\linewidth]{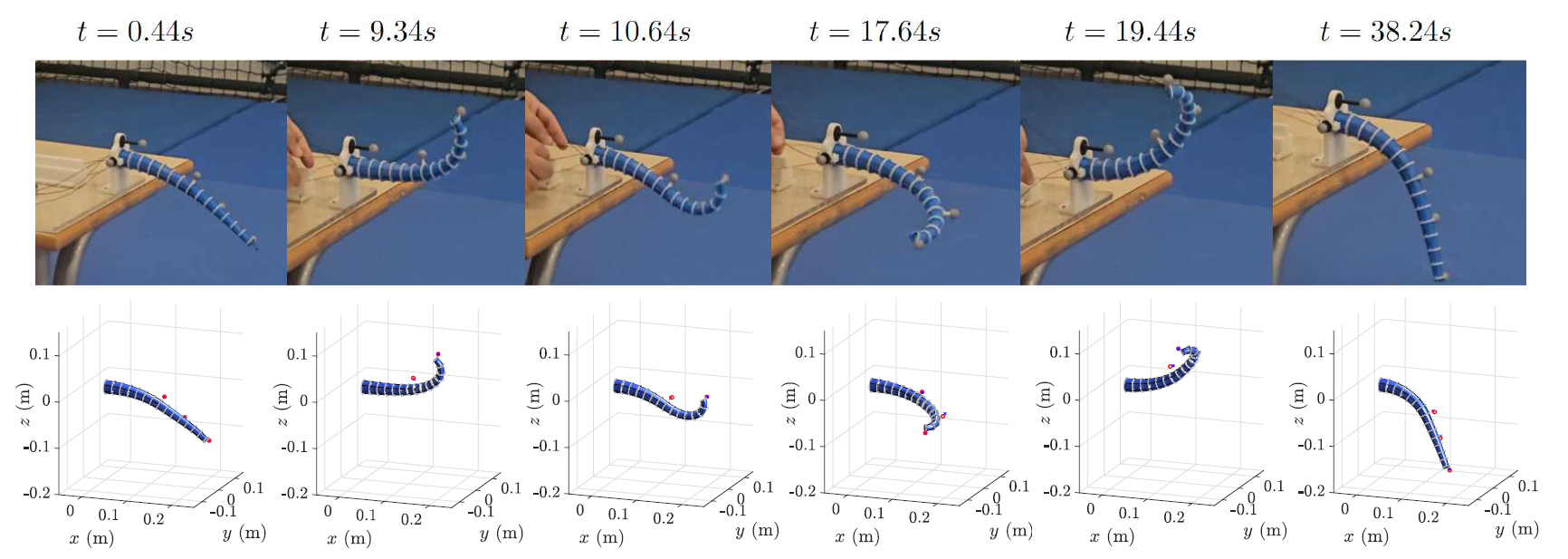}
    \caption{\small Comparison of the true shape and estimated shape using the experimental prototype. Motion tracking markers are shown in the second row of panels as the black x's, while their corresponding estimated counterparts are shown in red circles.}
    \label{fig::Exp_Shapes}
\end{figure*}

\begin{figure}[h!]
    \centering
    \includegraphics[width=\linewidth]{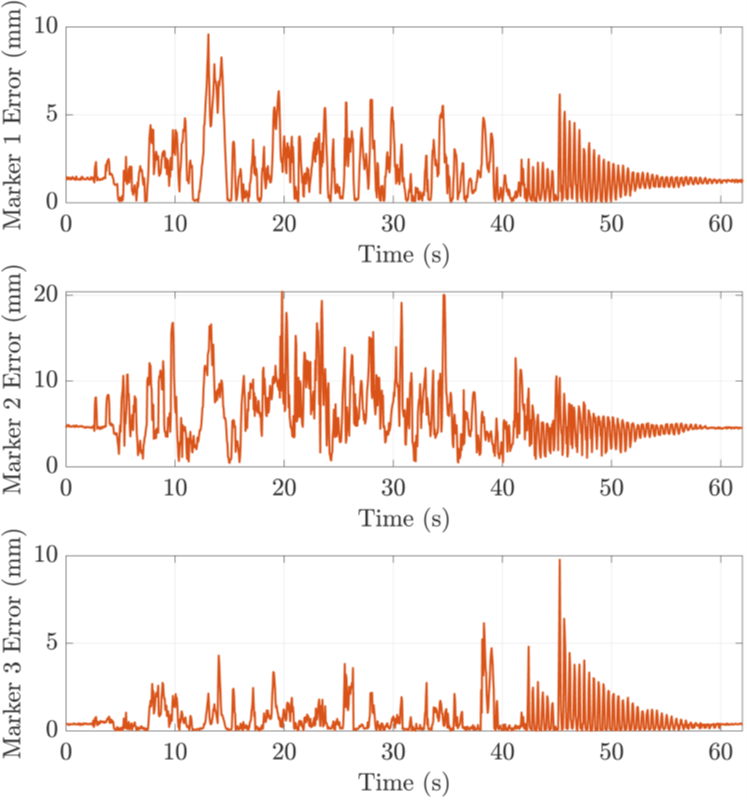}
    \caption{\small Errors for the three position markers. Marker 2, which used for evaluation exhibit higher errors than markers 1 and 3, whose errors are being minimized for the estimation problem. }
    \label{fig::Exp_Errors}
\end{figure}



The proposed approach is to estimate the generalized coordinates $\boldsymbol{q}^\mathfrak{e}$ that define the manipulator's shape such that the markers' positions match the measurements $\boldsymbol{p}^{\mathfrak{m}}$. Given the estimated generalized coordinates vector $\boldsymbol{q}^\mathfrak{e}$ with an optimal basis parametrization, the strain is:

\begin{equation}
\label{eq::estimatedstrain}
\bm{\xi}^\mathfrak{e}(X,\boldsymbol{q}^\mathfrak{e}) = {\boldsymbol{\Phi}_{\xi_\mathcal{O}}}(X)\boldsymbol{q}^{\mathfrak{e}}+\bm{\xi}^*,
\end{equation}
which can be integrated using equation (\ref{eq::int_kinem_pos}) to get the full robot configuration. The estimated transformation matrix of the $m^{th}$ marker can be expressed as follows:

\begin{equation}\label{fig::gei}
    \boldsymbol{g}^\mathfrak{e}_{m}(\boldsymbol{q}^\mathfrak{e}) = \boldsymbol{g}(X_{m},\boldsymbol{q}^\mathfrak{e}) \boldsymbol{g}_{m},
\end{equation}

\noindent where $X_{m}$ is the distance along the manipulator's centerline until the cross-section at which the $m^{th}$ marker is attached, and $\boldsymbol{g}_{m}$ is the known transformation between the centerline of the manipulator at $X_{m}$ and the marker position ($\boldsymbol{R}_{m} = \boldsymbol{I}$ and $\boldsymbol{r}_{m} = [0,\;p_{y_m},\;p_{z_m}]^T$). The estimated position of the markers $\boldsymbol{p}_{m}^\mathfrak{e}(\boldsymbol{q}^\mathfrak{e})$, can be obtained directly from $\boldsymbol{g}_{m}^\mathfrak{e} (\boldsymbol{q}^\mathfrak{e})$. We can define the error for the measurement in the $m^{th}$ marker as:

\begin{equation}    \boldsymbol{e}_{m} (\boldsymbol{q}^\mathfrak{e})={\boldsymbol{p}^\mathfrak{e}_{m} (\boldsymbol{q}^\mathfrak{e})-\boldsymbol{p}^{\mathfrak{m}}_{m}}
\end{equation}


Then, we define the following minimization problem:  

\begin{equation}\label{eq::ShapeEstOpt}
\begin{aligned}
\min_{\boldsymbol{q}^\mathfrak{e}} \quad & \lVert \boldsymbol{e}(\boldsymbol{q}^\mathfrak{e}) \lVert_{2}^{2}\\
\textrm{s.t.} \quad & \boldsymbol{q}_{max} > \boldsymbol{q}^\mathfrak{e} > \boldsymbol{q}_{min}
\end{aligned}
\end{equation}
where $\boldsymbol{e} = [\boldsymbol{e}_{1}^{T}, \boldsymbol{e}_{2}^{T},...,\boldsymbol{e}_{n_m}^{T}]^{T}$, $n_m$ is the number of markers used for the shape estimation and $\boldsymbol{q}_{max}$, $\boldsymbol{q}_{min}$ are the upper and lower bounds of $\boldsymbol{q}^\mathfrak{e}$, respectively.

The approach of estimating the shape with a few markers is made possible by the availability of the optimal bases obtained through the reduction approach discussed in Section \ref{sec::POD}. In Section \ref{sec::Analysis}, we demonstrated that the relative energy content $\varepsilon$ of the modes does not necessarily indicate the performance when included in the GVS model. However, this is not the case with the shape estimation problem. While including a few modes in the model could constrain the solution according to the allowed synergies, the shape estimation problem involves finding a linear combination of the chosen modes to minimize specific shape errors. In this sense, the shape estimation problem is similar to the SVD lower rank data reconstruction discussed in Section \ref{sec::truncation}. Consequently, $\varepsilon$ becomes a suitable metric for choosing the number of modes in this problem.

To find the optimal bases, we simulate 100 seconds of random actuation for our experimental prototype, similar to the approach presented in Section \ref{sec::6Act}. We then perform the POD of the resulting strain fields. Upon analyzing the decomposition results, we find that for the first 6 modes capture, $\varepsilon(6) = 99.65\%$, which we consider satisfactory. Consequently, the optimal bases are constructed from the leading 6 columns of the left singular matrix $\boldsymbol{U}$.

Another significantly important aspect of the decomposition is the insight we gain from the product of the singular values and the right singular matrix, $\boldsymbol{\Sigma V^T}$. While $\boldsymbol{U}$ provides the optimal modes in the data, $\boldsymbol{\Sigma V^T}$ reveals how these modes are scaled to obtain the different snapshots in the original data, whether it is a different instant of time or actuation value, as discussed earlier in Section \ref{sec::POD}. In other words, knowledge of $\boldsymbol{\Sigma V^T}$ is equivalent to knowledge of the generalized coordinates $\boldsymbol{q}$ of the data if it were generated by the optimal bases. Therefore, if the actuation range in the data is representative of the operating range, the limits of each mode, i.e., the maximum and minimum values of the generalized coordinates, $\boldsymbol{q}_{\max}$ and $\boldsymbol{q}_{\min}$, are known. This knowledge has been shown to significantly enhance the estimation process by providing bounds for the generalized coordinates during shape estimation from limited sensor measurements.

We use 2 position markers to estimate the 6 strain coefficients. We include an additional intermediate marker along the manipulator's body solely for evaluation purposes. To establish a position and orientation reference frame, three markers are placed at the base of the manipulator. These markers remain stationary throughout the experiment and serve as a fixed reference for the moving markers on the manipulator's body.

\begin{table}[h]
\centering
\caption{\small Marker positions for the experimental prototype used in the shape estimation problem. Marker 2 is used for evaluation, while markers 1 and 3 are used for the estimation problem.}

\begin{tabular}{cccc}

Marker index &
  \begin{tabular}[c]{@{}c@{}}$X_{m} (m)$\end{tabular} &
  \begin{tabular}[c]{@{}c@{}}$p_{y_m} (m)$\end{tabular} &
  \begin{tabular}[c]{@{}c@{}}$p_{z_m} (m)$\end{tabular} \\ \hline
1 & $0.125$ & $0$      & $0.025$          \\ \hline
2 & $0.18$ & $0.02$     & $0$          \\ \hline
3 & $0.25$ & $0.01$     & $0$          \\ \hline
\end{tabular}

\label{table::markerPositions}
\end{table}

\begin{figure*}[ht]
    \centering
    \includegraphics[width=\linewidth]{Figures/SLP_EXP_FIG_1.jpg}
    \caption{\small Comparison between experimental, HOM and ROM dynamic behaviour. Motion tracking position markers are shown in the simulation panels as the black x's, while their corresponding simulation counterparts are shown in red circles. The lines show the path of each marker over the preceding 0.1 seconds.}
    \label{fig::SLP_EXP}
\end{figure*}

During the experiment, the robot is actuated manually over 62 seconds, by pulling different tendons to obtain complex shapes and cover various areas within the robot's workspace. The marker positions are recorded using a motion capture system, which tracks the 3D coordinates of each marker over time. The resulting data consists of the time-varying positions of the markers along the manipulator's body, as well as the fixed reference markers at the base. This dataset will be used to validate the accuracy of our shape estimation algorithm and assess its performance in reconstructing the manipulator's shape based on the limited marker measurements.

Once the measurements are collected and preprocessed, we use the Levenberg-Marquardt nonlinear least squares algorithm, implemented through the MATLAB function \emph{lsqnonlin}, to solve the minimization problem in Eq. (\ref{eq::ShapeEstOpt}). We derive the error Jacobian of $\boldsymbol{e}(\boldsymbol{q}^\mathfrak{e})$, which can be used in the MATLAB function lsqnonlin to implement the Jacobian error of the optimization problem. This significantly enhances the efficiency of the optimization process towards achieving the optimal solution. The Jacobian of the components of $\boldsymbol{e}(\boldsymbol{q}^\mathfrak{e})$ can be computed as follows:


\begin{equation}
\begin{aligned}
 \frac{\partial  \boldsymbol{e}_{m}}{\partial \boldsymbol{q}^\mathfrak{e}} = & [\boldsymbol{\vartheta}^{\mathfrak{s}}_{m,1},\boldsymbol{\vartheta}^{\mathfrak{s}}_{m,2},...,\boldsymbol{\vartheta}^{\mathfrak{s}}_{m,n}] \\
 = & [\hat{\boldsymbol{J}}^{\mathfrak{s}}_{1}(X_{m})\overline{\boldsymbol{p}}^\mathfrak{e}_{m} ,\hat{\boldsymbol{J}}^{\mathfrak{s}}_{2}(X_{m})  \overline{\boldsymbol{p}}^\mathfrak{e}_{m},...,\hat{\boldsymbol{J}}^{\mathfrak{s}}_{n}(X_{m}) \overline{\boldsymbol{p}}^\mathfrak{e}_{m}]_3
\end{aligned}
\end{equation}

\noindent where $\overline{\boldsymbol{p}}^\mathfrak{e}_{m} = [(\boldsymbol{p}^\mathfrak{e}_{m})^{T}, 1]^{T}$,  $\boldsymbol{\vartheta}^{\mathfrak{s}}_{m,c}$ is the increment in the spatial velocity of the $m^{th}$ marker due to a small variation of the $c^{th}$ generalized coordinate, $\hat{\boldsymbol{J}}^{\mathfrak{s}}_{c}$ is the $c^{th}$ column of spatial Jacobian associated with the $c^{th}$ mode, and $[\bullet]_{3}$ extracts the first three rows of a homogeneous vector. The spatial Jacobian can be defined as $\boldsymbol{J}^{\mathfrak{s}}(X)= \mathrm{Ad}_{\bm{g}(X)} \boldsymbol{J}(X)$. 

In our experiment, we validate the real-time capabilities of our
shape estimation approach. When including the Jacobian, we achieve an average computation time of 34.3 ms, with a standard deviation of 14.1 ms for each sample. However, without including the Jacobian, the average computation time rises to 78.7 ms, with a standard deviation of 22.8 ms. Fig. \ref{fig::Exp_Shapes} displays the results of several shapes alongside their corresponding 3D reconstructions, demonstrating the effectiveness of our method in accurately estimating the robot's shape.

\begin{figure}[]
    \centering
    \includegraphics[width=\linewidth]{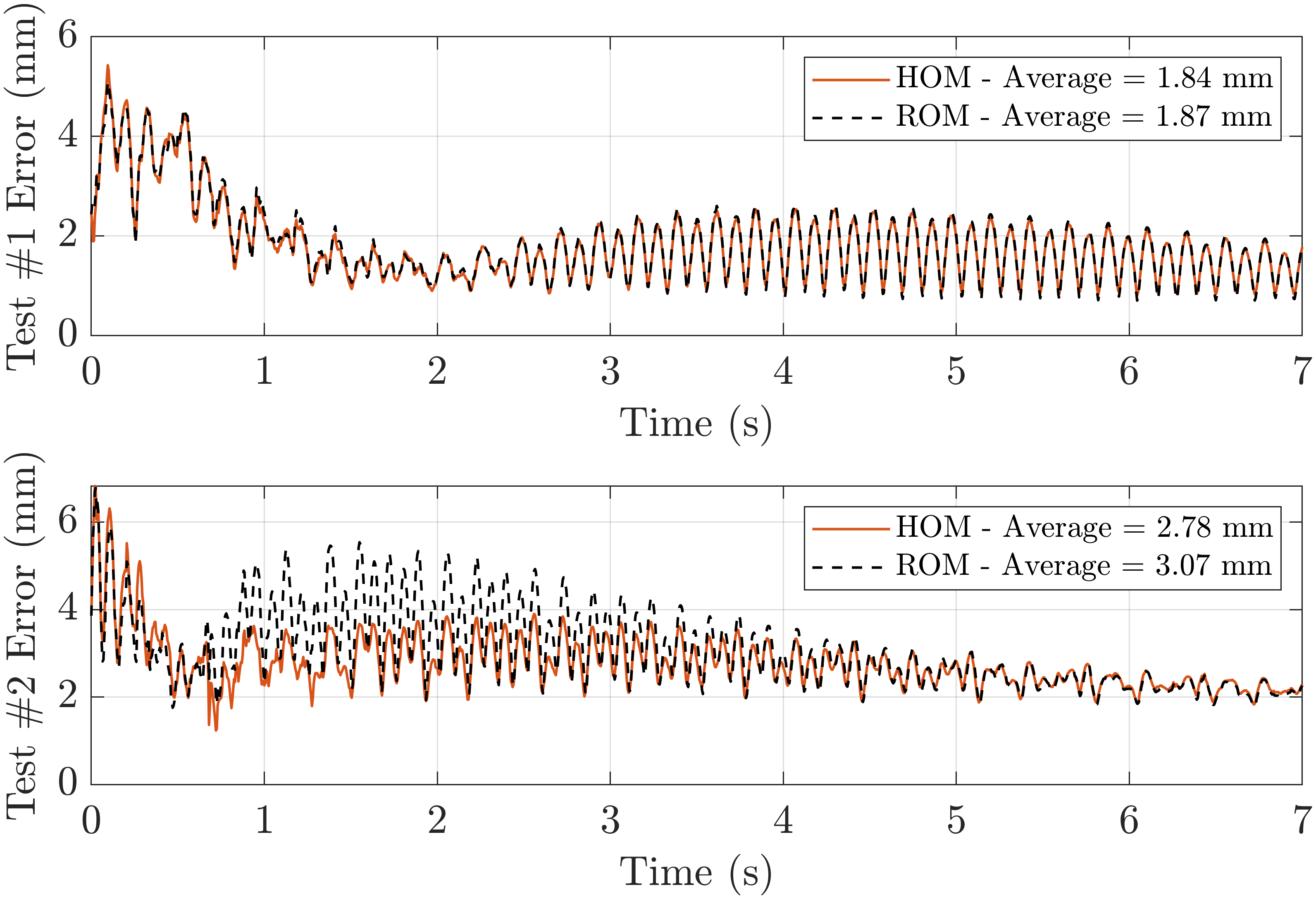}
    \caption{\small Average position errors for all markers over the first seven seconds of each test.} 
    \label{fig::SLP_Exp_Errors}
\end{figure}

To quantitatively evaluate the performance of our shape estimation approach, we analyze the position errors of the three markers throughout the full experiment, as shown in Fig. \ref{fig::Exp_Errors}. Markers 1 and 3, whose errors are being minimized in the optimization process, exhibit low errors as expected, with an average of 1.2 mm and a maximum of 10 mm. Marker 2, which is used for independent validation, also shows very small errors, with an average of 5.8 mm and a maximum of 20 mm. These results demonstrate the high accuracy and applicability of our method. While we restrict our experiment to use only two position markers and six modes for the shape estimation, using more modes and markers would further decrease the error.

The successful performance of our shape estimation approach in this example illustrates its potential to impact various other fields where accurate modeling and control of soft robots are crucial. The proposed method's ability to accurately reconstruct the robot's shape based on limited sensor data highlights its importance and opens up new possibilities for applications in diverse domains.


\subsection{ROM Validation}

To validate the ROM and assess its performance against reality, we conducted additional experiments on a different system. This system is a 1:2 scale model of the Soft-Legged Platform described in Section \ref{sec::SLP}, with fixed joints replacing the rotary joints. Seven position markers were placed on the moving parts of the system: one at the center of each soft leg and four on the rigid platform, as shown in Fig \ref{fig::SLP_EXP}. A preliminary set of experiments of dynamic behaviour were performed on the assembly to determine the material parameters. From these experiments, the density, Young's modulus, and damping coefficient were identified as 1185 $kg/m^3$, 0.69 MPa, and 350.4 Pa·s, respectively. We accounted for discrepancies in bending stiffness along the longer side (height) of the rectangular cross-section, with the stiffness being 1.67 times higher in that direction. This can be attributed to geometric uncertainties and the cubic relationship between the moment of inertia and cross-sectional height.

\subsubsection{Dynamic Behaviour}
We followed a methodology similar to previous examples to generate data and obtain the optimal basis. However, applying a controlled concentrated wrench directly to the platform, as done in Section \ref{sec::SLP}, proved impractical experimentally. Instead, we applied a wrench for one interval and removed it in the following interval, allowing the transient behavior and a decay towards equilibrium. Experimentally, this is equivalent to holding the platform in a deformed state and then releasing it. We simulated 5 minutes of system dynamics, with the platform loaded randomly and released every 1.5 seconds. The same strain parameterization as in Section \ref{sec::SLP} was used for the soft legs, resulting in a total of 54 DOFs.

After decomposition and acquiring the optimal basis set, we conducted a positional truncation analysis, similar to the one presented earlier in Fig. \ref{fig::DOF_VS_ErrTime_SLP}. We found that 21 DOFs resulted in a significant drop in position error, representing a clear truncation point. At this level of truncation, the speed-up factor was approximately 13.

During the experiment, we manually held the platform in a deformed configuration, simulating a constant wrench on the top rigid platform, and then released it instantly. Our goal was to compare the system's autonomous dynamics with both the ROM and HOM. To initiate the dynamic simulation, the initial state just before releasing the platform was required. For the ROM, we solved the shape estimation problem (\ref{eq::ShapeEstOpt}), presented earlier in Section \ref{sec::ShapeEstimation}, using the optimal strain modes and the markers' position measurements. Once the ROM's initial state was obtained, the corresponding state for the HOM could be calculated. Assuming that the optimal basis can reconstruct the entire strain field, i.e., \(\bar{\bar{\bm{\xi}}}_{H} = \bar{\bar{\bm{\xi}}}_{R}\) (subscripts denote whether it is the HOM or the ROM), the HOM state was computed as:
\begin{equation}
    \bm{q}_{H} = \left[ \begin{array}{ccc} \bm{\Phi}_{\xi} & \bm{0}  & \bm{0}  \\ \bm{0} & \bm{\Phi}_{\xi}  & \bm{0}  \\ \bm{0} & \bm{0}  & \bm{\Phi}_{\xi}  \end{array} \right]^\dag \bm{\Phi}_{\bar{\bar{\xi}}_{\mathcal{O}}} \bm{q}_{R}
\end{equation}
where here $\bm{\Phi}_{\xi} \in \mathbb{R}^{6\mathfrak{p} \times 18}$ is the strain basis of each soft leg, $\bm{\Phi}_{\bar{\bar{\xi}}_{\mathcal{O}}} \in \mathbb{R}^{18\mathfrak{p} \times 21}$ is the optimal basis matrix for the whole system (the three legs combined). and $\mathfrak{p}$ is the number of computational points per leg.

We conducted two tests, each starting from a different initial state, where the load was suddenly removed, and the system was allowed to return to equilibrium. Snapshots from both tests at various time points are shown in Fig. \ref{fig::SLP_EXP}, alongside equivalent moments from the HOM and ROM. Additionally, Fig. \ref{fig::SLP_Exp_Errors} illustrates the position error for each marker, with a maximum of approximately 6 mm in both tests. The average errors for the HOM and ROM in test 1 were 1.84 mm and 1.87 mm, respectively, and in test 2 were 2.78 mm and 3.07 mm, respectively. The minimal performance deterioration in the ROM demonstrates its ability to represent such a complex system with high accuracy.

\begin{figure}[]
    \centering
    \includegraphics[width=\linewidth]{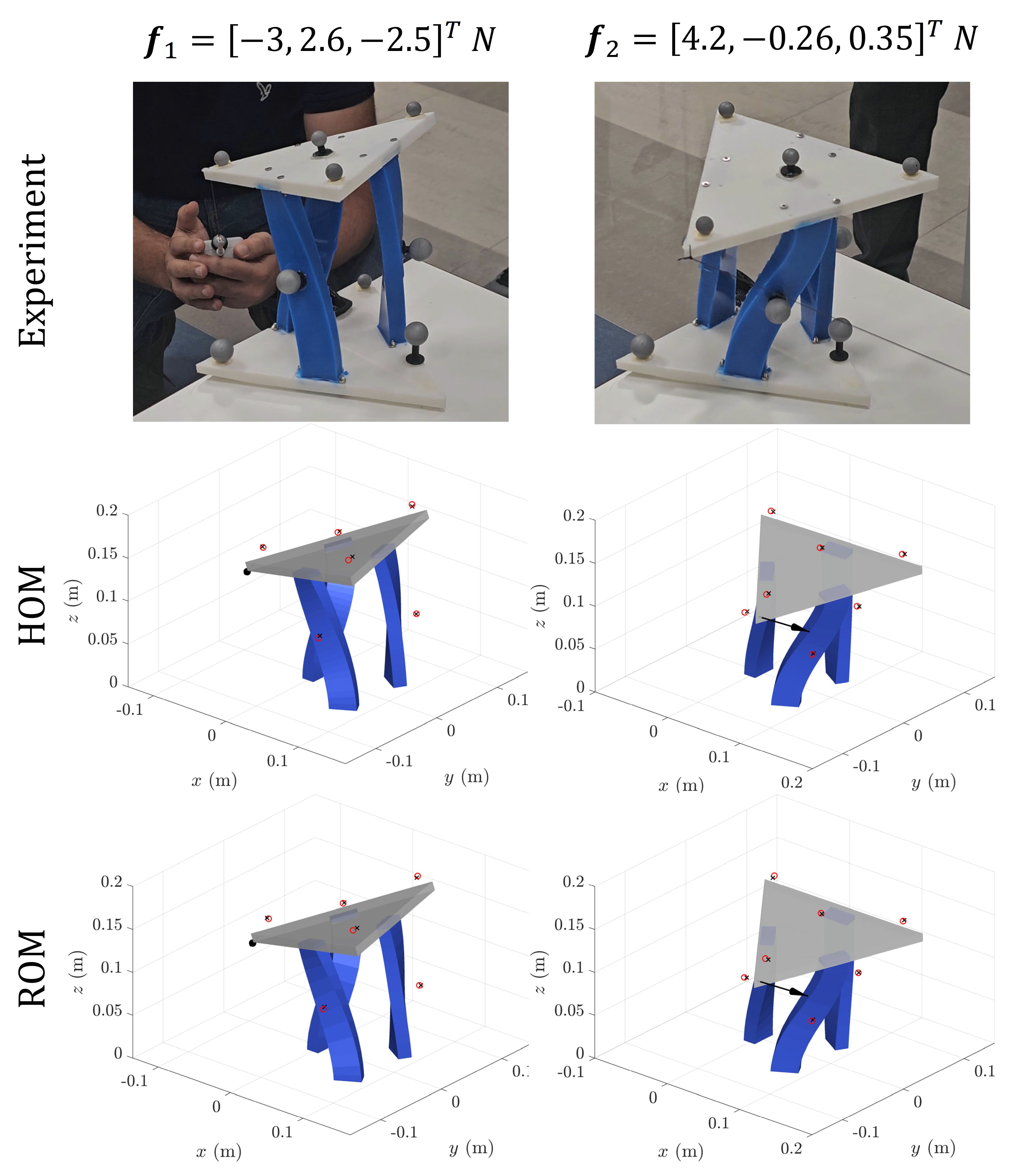}
    \caption{\small  Comparison between experimental, HOM, and ROM static behavior under a known constant load. Motion tracking markers are shown as black X's in the simulation panels, while their corresponding simulation counterparts are shown as red circles. The errors per marker for the HOM and ROM are 3.78 mm and 3.9 mm under $\bm{f}_1$, and 2.59 mm and 2.62 mm under  $\bm{f}_2$.} 
    \label{fig::SLP_Exp_Statics}
\end{figure}

\subsubsection{Static Behaviour}
We conducted another experiment to compare the static behavior of the system under a known constant load. In this test, an inextensible thread was connected to one edge of the platform, with the other end attached to a digital force gauge. A position marker was also placed on the force gauge to determine the direction of the applied force. Using the point of application and the force gauge marker, we calculated the applied wrench at the CM of the platform, which served as an input to the model.

To identify the optimal basis for this scenario, we applied POD to strain snapshots from 3000 random 3D forces exerted at the corresponding application point on the platform. The maximum applied force was 5 N, leading to significant system deformations. Similar to the dynamic analysis, we determined that 21 DOFs were sufficient to accurately represent the system.

We then compared the experimental results with those of the HOM and ROM under the measured loads. The system was subjected to two forces, $\bm{f}_1$ and $\bm{f}_2$, with different magnitudes and directions, expressed in the global frame. Fig. \ref{fig::SLP_Exp_Statics} shows the components of $\bm{f}_1$ and $\bm{f}_2$ and their corresponding behaviors in the simulation. For $\bm{f}_1$ , the errors per marker for the HOM and ROM were 3.78 mm and 3.9 mm, respectively, while for $\bm{f}_2$, they were 2.59 mm and 2.62 mm, respectively. Additionally, we observed a simulation speed-up factor of 2.1 at the chosen number of DOFs. Once again, the minimal increase in error for the ROM highlights its effectiveness in replicating the HOM's performance.

\section{Discussion and Conclusion}

This work introduced a novel POD-based reduced order modeling approach for soft and hybrid robots combined with the GVS formulation. By learning optimal strain basis functions from data, the method allows a minimal set of generalized coordinates to efficiently describe robot configurations. Extensive simulations and experiments validated the accuracy and computational benefits of the ROM across diverse soft robotic systems. Notably, the ROM enabled shape estimation for a physical soft manipulator using only two position markers, demonstrating its practical value.

The proposed approach offers several key advantages. First, it achieves substantial dimensionality reduction without compromising accuracy. This resulted in a significant reduction in the computational cost of simulations, as the system dimension is one of the main factors governing the computational cost as seen in Figs. \ref{fig::DOF_VS_ErrTime}, \ref{fig::DOF_VS_ErrTime_SLP}, and \ref{fig::SpiralSnapshots}. Additionally, in Table \ref{table::SpeedupTable}, we report the speed-up factors that correspond to ROMs exhibiting 5\% tip position errors for our presented scenarios. Notably, we achieve speed-up factors exceeding 16 while maintaining acceptable performance. The computational efficiency of our model allows simulations to run faster than real-time on an average computer (in specific cases), paving the way for real-time applications. For example, in the scenario depicted in Fig. \ref{fig::SingleCab_modes_DynWGrav}, we computed 10 seconds of dynamics under continuously changing actuation within just 2 seconds, utilizing all four main modes. This demonstrates our approach's suitability for real-time applications in specific cases, such as control and estimation. Second, it provides physically interpretable soft synergies in the form of coupled strain modes. These modes provide insights on the system behaviour, showcasing the dominant features in the system, and their corresponding contributions to the solution, as seen in Figs. \ref{fig::SingleCabNoGrav_mode},\ref{fig::SingleCab_modes_DynWGrav}, and \ref{fig::6Cables_Mode}. Such understanding of the system allowed us to estimate the entire shape of a complex soft manipulator, using only the position of two markers along its body. Third, the ROM exhibits strong interpolation and extrapolation capabilities, facilitating its use within and beyond the original data range. Finally, the unified treatment of strains and joint twists allows seamless application to entirely rigid or hybrid rigid-soft systems.

Future work will explore further applications of the ROM, such as model-based control, motion planning, and contact modeling for soft robots. In addition, incorporating modal derivatives \cite{wu_modal_2019}, i.e., how the modes change according to the system state, offers a promising way to introduce further enhancement to the reduction and accuracy of our approach. Also, using deep-learning methods to reduce a non-linearly strain-parameterized model holds potential to provide further reduction to the system dimensionality, in contrast to linear strain bases presented in this work.  Furthermore, improvement in other factors that affect the computation time, such as time integration methods, hold the potential of making such simulations faster and more reliable. With growing interest in soft robotics for safe and adaptable human-robot interaction, ROMs enhanced with data-driven approaches will play an increasingly important role in their modeling, control, estimation, etc. The approach presented here offers a promising foundation for realizing the full potential of soft robots in real-world applications.

\section*{Acknowledgments}

This work was supported by the US Office of Naval Research Global under Grant N62909-21-1-2033 and in part by Khalifa University under Awards No. RIG-2023-048,
RC1-2018-KUCARS.


%

\appendices
\section{Multimedia Appendix}
The appendix features supplementary video material that showcases the performance of our proposed approach compared to HOMs including static and dynamic simulations of the presented prototypes using varying numbers of DOFs. Furthermore, the video presents a side-by-side comparison of the real prototype from the shape estimation experiment and the shape estimated using our method, highlighting the high qualitative performance of the estimation. It also includes a comparison between the real system, HOM and ROM for the static and dynamic validation experiments conducted on the parallel hybrid soft-rigid system.

\section{Coefficients of generalized dynamics}

\label{app:B}
For a hybrid multibody system with $N$ links,
\begin{subequations}
\begin{align}
&\bm{M}\left(\bm{q}\right) = \sum_{i=1}^N \int_0^{L_i} \bm{J}_i^T \overbar{\bm{\mathcal{M}}}\bm{J}_i dX_i  \\
&\bm{C}\left(\bm{q},\dot{\bm{q}}\right) = \sum_{i=1}^N \int_0^{L_i} \bm{J}_i^T \left( \mathrm{ad}_{\bm{\eta}_i}^*\overbar{\bm{\mathcal{M}}}\bm{J}_i + \overbar{\bm{\mathcal{M}}}\dot{\bm{J}}_{i} \right) dX_i \\
&\bm{D} = \text{diag}_{i=1}^N \left( \int_0^{L_i} \bm{\Phi}_{\xi_i}^T \bm{\Upsilon}_i \bm{\Phi}_{\xi_i} dX_i \right)   \\
&\bm{K} = \text{diag}_{i=1}^N \left( \int_0^{L_i} \bm{\Phi}_{\xi_i}^T \bm{\Lambda}_i \bm{\Phi}_{\xi_i} dX_i \right)   \\
&\bm{B}\left(\bm{q}\right) = \text{diag}_{i=1}^N \left( \bm{S}_{i}^T \text{ or } \int_0^{L_i} \bm{\Phi}_{\xi_i}^T \bm{\Phi}_{a_i} dX_i   \right) \\
&\bm{F}\left(\bm{q}, \dot{\bm{q}},t \right) = \sum_{i=1}^N \int_0^{L_i} \bm{J}_i^T \overbar{\bm{\mathcal{F}}}_{e_i} dX_i 
\end{align}
\end{subequations}
where, $\overbar{\bm{\mathcal{M}}}$ is the screw inertia density matrix, $\bm{\Upsilon}$ is the screw damping matrix, $\bm{\Lambda}$ is the screw elasticity matrix, $\bm{S}_{i}$ is the joint actuation basis, and $\overbar{\bm{\mathcal{F}}}_{e_i}$ is the distributed external force like gravity. Note that the space integration is evaluated using a numerical integration scheme such as Gauss quadrature:
\begin{equation}
\int_{0}^{L_i} \bm{f}(X) \,dX
 =  \sum_{j=1}^{Q_i} w_{j} \bm{f}(X_{j})
\end{equation}
where $Q_i$ is the number of quadrature points on $i^{th}$ link, $w_j$ is the $j^{th}$ quadrature weight.

The equation also applies to rigid bodies by eliminating the integrals and substituting the distributed quantities with their lumped equivalents. The generalized external forces due to point wrenches are computed by projecting them with the Jacobian at the points of application of the wrenches: $\bm{J}^T \bm{\mathcal{F}}_{p}$.

For the reduced order basis, $\bm{D}$, $\bm{K}$, and $\bm{B}$ are modified into:
\begin{subequations}
\begin{align}
&\bm{D} = \sum_{i=1}^N  \sum_{j=1}^{Q_i} w_{j} (\bm{\mathcal{E}}_k{\boldsymbol{\Phi}_{\bar{\bar{\xi}}_\mathcal{O}}})^T \bm{\Upsilon}_j (\bm{\mathcal{E}}_k{\boldsymbol{\Phi}_{\bar{\bar{\xi}}_\mathcal{O}}})    \\
&\bm{K} = \sum_{i=1}^N  \sum_{j=1}^{Q_i} w_{j} (\bm{\mathcal{E}}_k{\boldsymbol{\Phi}_{\bar{\bar{\xi}}_\mathcal{O}}})^T \bm{\Lambda}_j (\bm{\mathcal{E}}_k{\boldsymbol{\Phi}_{\bar{\bar{\xi}}_\mathcal{O}}})    \\
&\bm{B}\left(\bm{q}\right) = \sum_{i=1}^N \left( \bm{S}_{i}^T \text{ or } \sum_{j=1}^{Q_i} w_{j} (\bm{\mathcal{E}}_k{\boldsymbol{\Phi}_{\bar{\bar{\xi}}_\mathcal{O}}})^T \bm{\Phi}_{a_j}   \right) 
\end{align}
\end{subequations}
where $k$ is the index of the computational point of the multibody system corresponding to the $j^{th}$ quadrature point of the link.

\ifCLASSOPTIONcaptionsoff
  \newpage
\fi



%
%



\bibliographystyle{ieeetr}
\bibliography{arxiv}




\end{document}